\documentclass[lettersize,journal]{IEEEtran}
\usepackage{amsmath,amsfonts}
\usepackage{algorithmic}
\usepackage{algorithm}
\usepackage{array}
\usepackage[caption=false,font=normalsize,labelfont=sf,textfont=sf]{subfig}
\usepackage{textcomp}
\usepackage{stfloats}
\usepackage{url}
\usepackage{verbatim}
\usepackage{graphicx}
\usepackage{cite}
\usepackage{color} 
\usepackage{verbatim} 
\usepackage{amsfonts} 
\usepackage{amssymb} 
\usepackage{bbding} 
\usepackage{multirow} 
\usepackage{multicol} 
\usepackage{makecell} 
\usepackage{pifont} 

\usepackage{hyperref}
\hypersetup{hypertex=true,
            colorlinks=true,
            linkcolor=blue,
            anchorcolor=blue,
            citecolor=blue,
            urlcolor=black}

\usepackage{booktabs} 
\usepackage[table]{xcolor}

\usepackage{comment} 

\usepackage{eso-pic}
\usepackage{xcolor}

\AddToShipoutPicture*{
    \put(40,20){
    \fcolorbox{black}{white}{
      \parbox{\textwidth}{
        \color{red}\footnotesize
        \textcopyright~2025 IEEE. Personal use of this material is permitted. Permission from IEEE must be obtained for all other uses, in any current or future media, including reprinting/republishing this material for advertising or promotional purposes, creating new collective works, for resale or redistribution to servers or lists, or reuse of any copyrighted component of this work in other works.
      }
    }
    }
}

\begin{document}



\title{Ambiguity-aware Point Cloud Segmentation by Adaptive Margin Contrastive Learning}

\author{Yang Chen, Yueqi Duan,~\IEEEmembership{Member,~IEEE}, Haowen Sun, Jiwen Lu,~\IEEEmembership{Fellow,~IEEE}, \\ and Yap-Peng Tan,~\IEEEmembership{Fellow,~IEEE}
\thanks{This work is supported in part by the National Natural Science Foundation of China under Grant 62206147 and in part by the National Research Foundation of Singapore under the NRF Medium Sized Centre Scheme (CARTIN). \textit{(Corresponding author: Yueqi Duan.)}

Yang Chen and Yap-Peng Tan are with Nanyang Technological University, Singapore (e-mail: chen1605@e.ntu.edu.sg; eyptan@ntu.edu.sg). Yueqi Duan, Haowen Sun, and Jiwen Lu are with Tsinghua University, China (e-mail: duanyueqi@tsinghua.edu.cn; sunhw24@mails.tsinghua.edu.cn; lujiwen@tsinghua.edu.cn).}
}


\markboth{THIS ARTICLE HAS BEEN ACCEPTED FOR PUBLICATION IN IEEE TRANSACTIONS ON MULTIMEDIA. DOI:XX.XXX/XXX.XX.XXXX}
{Shell \MakeLowercase{\textit{et al.}}: A Sample Article Using IEEEtran.cls for IEEE Journals}



\maketitle

\begin{abstract}

This paper proposes an adaptive margin contrastive learning method for 3D semantic segmentation on point clouds. Most existing methods use equally penalized objectives, which ignore the per-point ambiguities and less discriminated features stemming from transition regions. However, as highly ambiguous points may be indistinguishable even for humans, their manually annotated labels are less reliable, and hard constraints over these points would lead to sub-optimal models. To address this, we first design AMContrast3D, a method comprising contrastive learning into an ambiguity estimation framework, tailored to adaptive objectives for individual points based on ambiguity levels. As a result, our method promotes model training, which ensures the correctness of low-ambiguity points while allowing mistakes for high-ambiguity points. As ambiguities are formulated based on position discrepancies across labels, optimization during inference is constrained by the assumption that all unlabeled points are uniformly unambiguous, lacking ambiguity awareness. Inspired by the insight of joint training, we further propose AMContrast3D++ integrating with two branches trained in parallel, where a novel ambiguity prediction module concurrently learns point ambiguities from generated embeddings. To this end, we design a masked refinement mechanism that leverages predicted ambiguities to enable the ambiguous embeddings to be more reliable, thereby boosting segmentation performance and enhancing robustness. Experimental results on 3D indoor scene datasets, S3DIS and ScanNet, demonstrate the effectiveness of the proposed method. Code is available at \href{https://github.com/YangChenApril/AMContrast3D}{\textit{https://github.com/YangChenApril/AMContrast3D}}.

\end{abstract}


\section{Introduction}
\label{sec:intro}

3D semantic segmentation is a fundamental task in generating semantic coherent regions~\cite{qi2017pointnet},~\cite{qi2017pointnet++},~\cite{thomas2019kpconv}, where point cloud is a common 3D representation that attracts increasing attention. Processing and segmenting individual points require local semantics with enhanced features for deep learning~\cite{kang2023region},~\cite{sun2024mirageroom},~\cite{thomas2019kpconv},~\cite{zhang2024geoauxnet},~\cite{zheng2024point},~\cite{song2022kernel},~\cite{fan2024look},~\cite{song2022lslpct}. Conventional deep learning approaches employ the cross-entropy objective to guide model training~\cite{qi2017pointnet},~\cite{qi2017pointnet++},~\cite{hu2020jsenet},~\cite{qian2022pointnext},~\cite{zheng2023learning}. Recently, more scholarly efforts have extended this paradigm by incorporating point-level contrastive objective~\cite{xie2020pointcontrast},~\cite{ tang2022contrastive},~\cite{li2022hybridcr},~\cite{jiang2021guided}, which serves as a complementary feature learning strategy to promote compactness within the same semantic regions and dispersion among different semantic regions.

Despite the effectiveness in enhancing feature discrimination, most prevailing contrastive objectives develop a uniform training difficulty for different points. However, points in transition regions, which commonly interconnect several semantic classes, often exhibit higher sparsity and irregularity compared to those points near the object centroid. This inherent disparity introduces inevitable per-point ambiguities that are challenging for both models and human annotators to distinguish. Consequently, when applying a uniform training difficulty to points in transition regions, the model unavoidably over-prioritizes the segmentation of these points and the optimization of their less discriminated features. This, in turn, results in a lack of attention towards the remaining crucial points, further leading to instability during model training.

\begin{figure}[t]
\begin{minipage}[b]{1.0\linewidth}
  \centering
  \includegraphics[width=8.8cm]{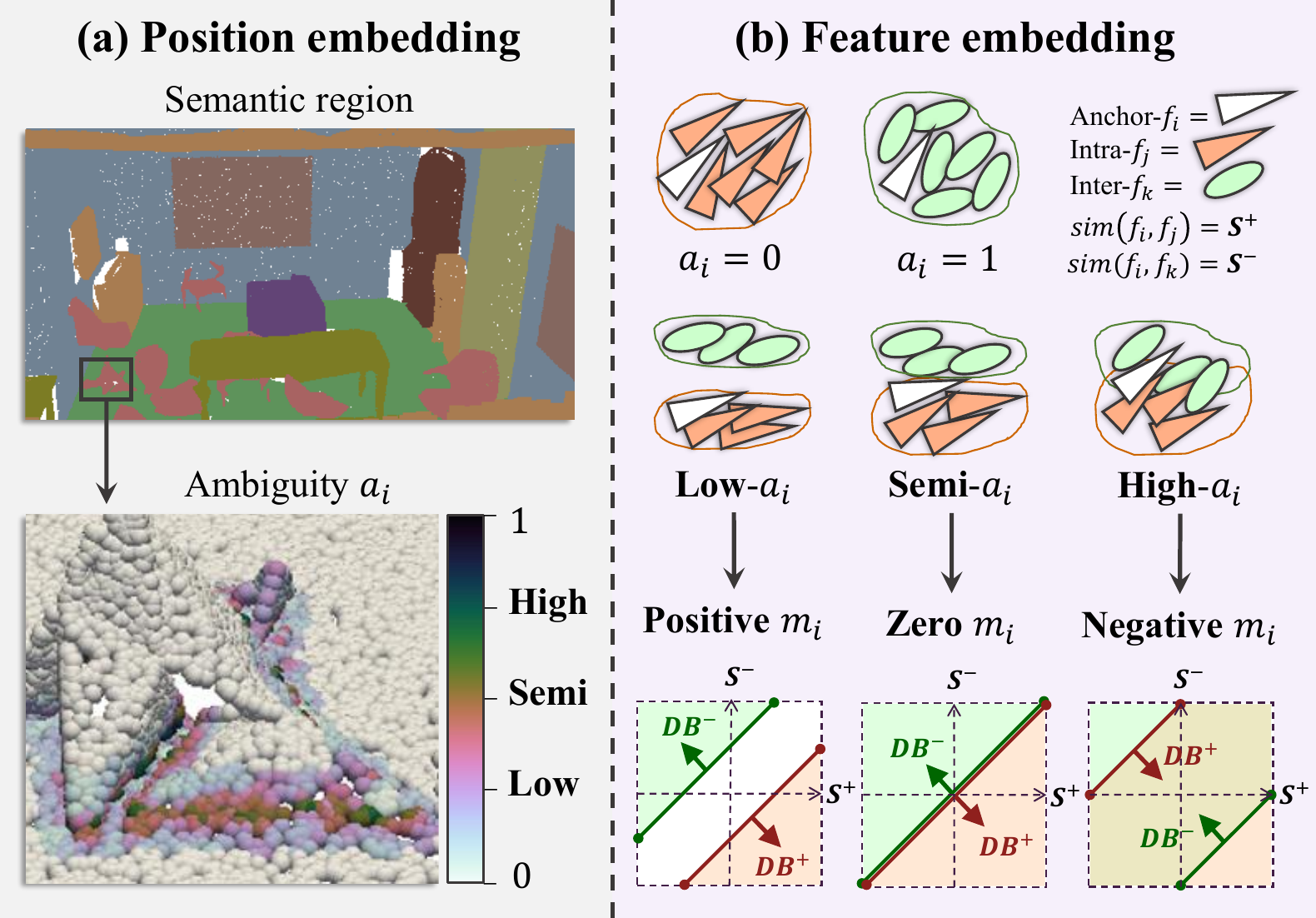}
\end{minipage}
\caption{Adaptive correspondence between ambiguity and margin. An illustration among (a) position embedding indicates per-point ambiguity $a_i$ colored by a map ranging from $0$ to $1$, and (b) feature embedding yields similarities of intra-pair $S^+$ and inter-pair $S^-$, using ambiguity-aware margin $m_i$ to adjust decision boundaries $DB^+$ and $DB^-$ in contrastive learning, which generates adaptive objectives to benefit embedding learning.}
\label{fig:teaser}
\end{figure}

Motivated by the disparity of per-point ambiguities, we introduce adaptive objectives tailored for training difficulties on different points. Aligning with 2D tasks that leverage decision margins to heighten training difficulty~\cite{deng2019arcface},~\cite{wang2018cosface},~\cite{li2022towards},~\cite{li2020boosting}, our method similarly preserves large margins for points with low ambiguities, yet narrows the margins to be smaller even down to negative values for highly ambiguous points in 3D space. In this way, margins positively correlate with training difficulties, and Fig.~\ref{fig:teaser} demonstrates that low-ambiguity, semi-ambiguity, and high-ambiguity points correspond to positive, zero, and negative margins between decision boundaries. Our method, namely \textbf{AMContrast3D} (Section~\ref{sec:method_AAA}), comprises an ambiguity estimation framework (AEF), from which we first compute per-point ambiguities from position embeddings, and points tightly surrounded by neighboring points of different semantic labels reflect high ambiguities. Furthermore, we construct an ambiguity-aware margin generator (MG), which dynamically integrates contrastive learning to regularize feature embeddings among intra-class and inter-class. By adjusting the adaptive margins in contrastive objectives, points in transition regions are effectively regularized, thereby strengthening the discrimination of features.

Although our method draws point ambiguity awareness by endowing contrastive objectives, the per-point ambiguities are formulated relying on annotated labels. As labels are not used for inference, ambiguity estimation becomes unfeasible after training, thereby leading to incomprehensive awareness of ambiguity. Consequently, points are uniformly processed with an unambiguous assumption, which imposes unreliable predictions for high-ambiguity points. Based on these observations, we further propose \textbf{AMContrast3D++} (Section~\ref{sec:method_MMM}) through an ambiguity awareness branch. Specifically, we first design a lightweight ambiguity prediction module (APM) to encode position and feature embeddings. Supervised by the point ambiguities formulated from AEF, our APM learns to output predicted ambiguities, from which these two types of ambiguities are regarded as supervision signals and regression predictions, respectively. To compensate for the limited ambiguity-aware optimization, we construct a masked refinement (MR) mechanism, where APM simultaneously enables MR to generate per-point masks and refine embeddings within neighboring regions, filtering out anchor embeddings of high-ambiguity points and extracting neighboring embeddings of low-ambiguity points. Equipped with well-designed modules and objectives, the ambiguity awareness branch is trained jointly under the semantic segmentation branch. This two-branch method develops the interactions between point ambiguities and related embeddings to enhance training and inference. Extensive experiments on S3DIS~\cite{armeni2017joint} and ScanNet~\cite{dai2017scannet} demonstrate that our methods outperform many methods. 

This paper is an extended version of our conference paper~\cite{chen2024adaptive}, where we make the following new contributions: 

\begin{enumerate}
\item We propose a new attributed method, AMContrast3D++, extending from the AMContrast3D in our conference version. We design an ambiguity awareness branch on the semantic segmentation branch, which jointly predicts point ambiguities and labels to realize a comprehensive optimization.

\item We design a lightweight ambiguity prediction module, which is under supervision and regularization. In addition, we propose a masked refinement mechanism by leveraging the prediction as auxiliary information, so highly ambiguous embeddings are detected and refined during training and inference.

\item We conduct quantitative experiments with ablations to present our leading performances. For reasonable evaluations, we perform extensive random runs to verify robustness through the highest and average results with error bounds. Moreover, we qualitatively show per-point ambiguity visualizations for challenging scenes.

\end{enumerate}

\section{Related Work}
\label{sec:rw}

\subsection{3D Point Cloud Segmentation}

PointNet~\cite{qi2017pointnet} pioneers the 3D segmentation, which directly works on irregular point clouds. This network processes individual points with shared MLPs to aggregate global features. However, its performance is limited because of the lack of considering local spatial relations in the point cloud structure. Furthermore, PointNet++~\cite{qi2017pointnet++} introduces a hierarchical spatial structure to aggregate features in local regions.

To improve feature representations in 3D point cloud scenes, recent studies have deep explorations and present advancements~\cite{song2022kernel},~\cite{fan2024look},~\cite{song2022lslpct}. To demonstrate improvements in local and global geometric representations, KCB~\cite{song2022kernel} adaptively captures features through kernel correlation. PS~\cite{fan2024look} presents a patch-wise saliency map, which emphasizes the importance of points to represent specific local structures. LSLPCT~\cite{song2022lslpct} parallelizes the perception of global contextual information and captures local semantic features, developing the essential feature precision on point clouds. These state-of-the-art studies address challenges in feature learning, and their developments integrate insights to handle uncertainty in complex 3D scenes, which are relevant for investigating ambiguous regions.

In MLPs-based philosophy, existing methods develop novel modules~\cite{liu2020closer},~\cite{qian2022pointnext},~\cite{lin2023meta}. PointNeXt~\cite{qian2022pointnext} revisits training and scaling strategies, tweaking PointNet++ based on the set abstraction block. The recently proposed method, PointMetaBase~\cite{lin2023meta}, designs building blocks into four meta functions for point cloud analysis. Compared with convolutional kernels~\cite{thomas2019kpconv},~\cite{xu2021paconv},~\cite{liu2020semantic}, graph structures~\cite{landrieu2018large},~\cite{qian2021pu},~\cite{chen2020hapgn},~\cite{tao2022seggroup}, and transformer architectures~\cite{zhao2021point},~\cite{park2022fast},~\cite{kang2023region}, the highly-optimized MLPs are conceptually simpler to reduce computational and memory costs and achieve better results than many complicated methods.

\subsection{Training Objective for Deep Learning} 

Contrastive learning methods~\cite{gutmann2010noise},~\cite{ oord2018representation},~\cite{khosla2020supervised} are used to pull together feature embeddings from the same class and push away feature embeddings from different classes. Works that follow this path design various contrastive objectives on 3D tasks in unsupervised approach~\cite{xie2020pointcontrast,wen20233d}, weakly-supervised approach~\cite{li2022hybridcr}, semi-supervised approach~\cite{jiang2021guided}, supervised approach~\cite{tang2022contrastive}, and self-supervised approach~\cite{tang2023point},~\cite{wu2023self},~\cite{liu2023inter}. However, they only conduct fixed contrast on feature embeddings while ignoring adaptive ambiguities from position embeddings. Besides, contrastive learning mechanisms can be applied to a wide range of fields, such as object detection~\cite{huang2024m}, image dehazing~\cite{cheng2024continual}, and recommendation system~\cite{ong2023quad}. M-RRFS~\cite{huang2024m} introduces a memory-based synthesizer to generate robust features for unseen categories, conducting zero-shot learning based on the contrast. Additionally, this work leverages the contrastive mechanism with region-level information, while our approach adjusts contrastive embeddings based on point-level features. KR~\cite{cheng2024continual} designs effective knowledge replay to conduct contrastive regularization, realizing feature enhancements and improving robustness across adverse weather images. These studies demonstrate novel perspectives to address various challenges in contrastive learning. 

Moreover, decision margin is witnessed as a method to adjust the objective and strengthen the discriminating power~\cite{deng2019arcface, wang2018cosface}. While 2D tasks~\cite{li2022towards, li2020boosting} have shown some success in adopting dynamic margins to enhance constant margins, they are mostly constrained to positive margins to heighten objectives. This direction is essentially an under-explored aspect of 3D tasks. Meanwhile, such positive margins are only aware of a one-sided situation that increases the training difficulties, which is restrictive due to the intrinsic properties of point clouds. In contrast, our method deviates from the one-sided methods, from which we design adaptive objectives to explore ambiguity-aware margins with diverse values.

\begin{figure*}[!t]
\centering
\includegraphics[width=7.0in]{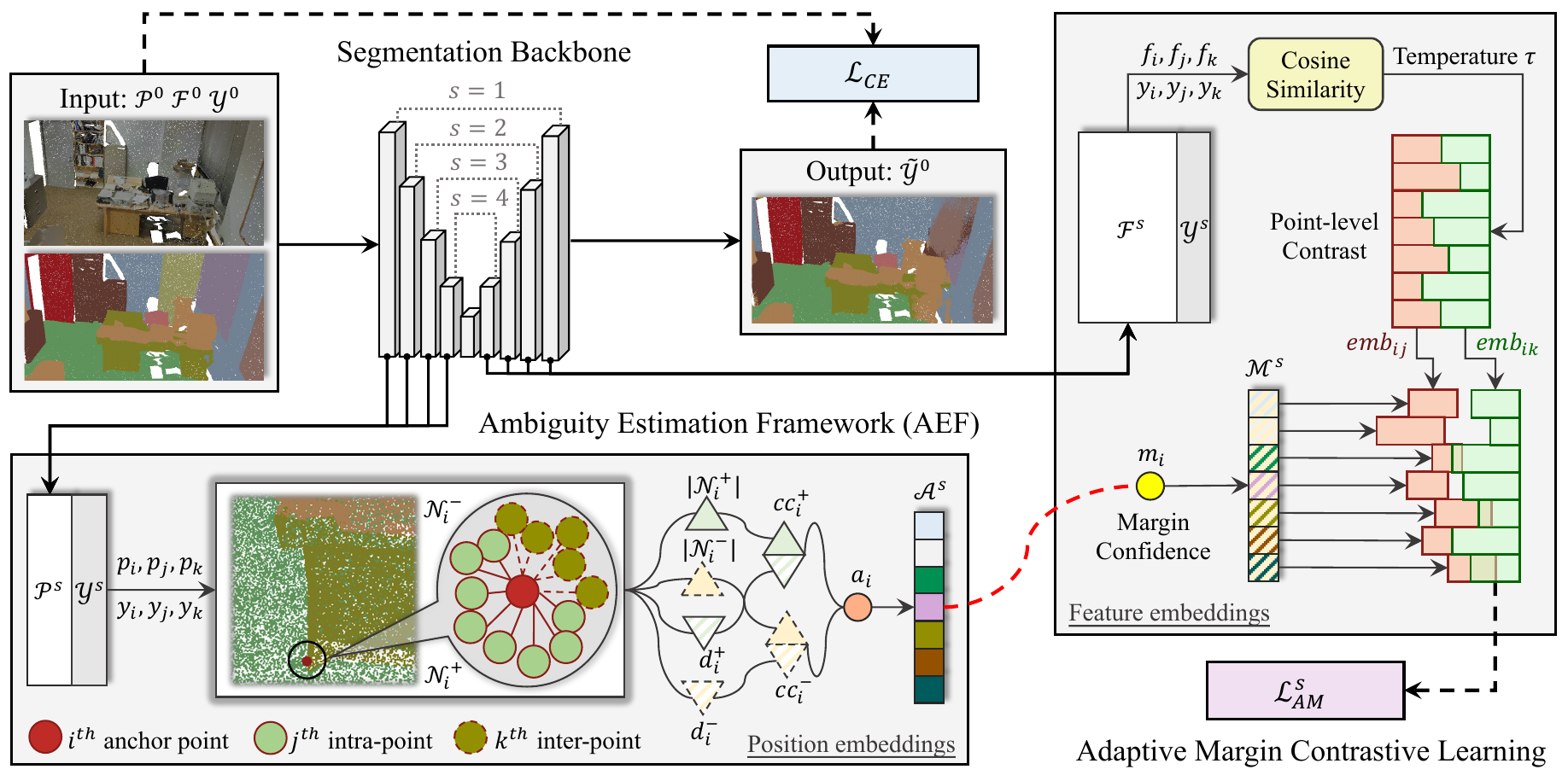}
\caption{The AMContrast3D architecture with segmentation backbone. In the ambiguity estimation framework following the encoding stage $s$, we infer the ambiguity $a_i \in \mathcal{A}^s$ for each $i^{th}$ point by encoding position embeddings $p_i,p_j, p_k \in \mathcal{P}^s$, where the $j^{th}$ intra-point in $\mathcal{N}_i^+$ and the $k^{th}$ inter-point in $\mathcal{N}_i^-$ are extracted through sub-scene labels $y_i,y_j,y_k \in \mathcal{Y}^s$. We reformulate $a_i$ into adaptive ambiguity-aware margins $m_i \in \mathcal{M}^s$. These margins target feature embeddings $f_i,f_j,f_k \in \mathcal{F}^s$ for corresponding decoding stages to dynamically adjust decision boundaries during contrastive learning. Through the adaptive margin contrastive learning, our method automatically regulates training difficulties across different parts of the point clouds, particularly ensuring more stabilized training for high-ambiguity points in transition regions containing different semantic classes.}
\label{fig:framework}
\end{figure*}

Another direction of training objective lies in multi-level supervision~\cite{gong2021omni},~\cite{tang2022contrastive}, where Omni-scale supervision~\cite{gong2021omni} applies MLPs to generate the prediction field from the target field at multiple stages. Joint training strategies integrate multiple objectives~\cite{weng2022context}. PU-Net~\cite{yu2018pu} combines reconstruction and repulsion objectives. JSENet~\cite{hu2020jsenet} explores boundary supervision for tasks with close relationships in the feature domain. Different from these works, we design an innovative task involving the awareness of 3D point ambiguities at all stages, benefiting embeddings and refining semantic predictions.

\section{AMContrast3D}
\label{sec:method_AAA}

\subsection{Problem Formulation}

Categorizing points to correct semantic classes is the goal of 3D semantic segmentation. Suppose $n$ is the number of points and $C$ is the number of classes in a point cloud, points are parsed to predict labels $\mathcal{\tilde{Y}}^0 = \{\tilde{y}_i\}_{i=1}^{n}$ aligning with the ground-truth labels $\mathcal{Y}^0 = \{y_i\}_{i=1}^{n}$, where $\tilde{y}_i, y_i \in \mathbb{R}^C$. Given point position $\mathcal{P}^0 = \{p_i\}_{i=1}^{n}$ and point feature $\mathcal{F}^0 = \{f_i\}_{i=1}^{n}$, each $i^{th}$ point has $p_i \in \mathbb{R}^{3}$ and $f_i \in \mathbb{R}^{D}$ with $D$ as feature dimension. Segmentation backbone directly ingests $\mathcal{P}^0$ and $\mathcal{F}^0$, stacking multiple encoding-decoding stages to learn underlying representations of individual points. We index each stage as a notation $s$ and re-sample $n^s$ points by the farthest point sampling algorithm. Each stage has $3$-dim $\mathcal{P}^s$ as position embeddings, $D^s$-dim $\mathcal{F}^s$ as feature embeddings, and $C$-dim $\mathcal{Y}^s$ as sub-label embeddings based on label mining strategy. The cross-entropy objective $\mathcal{L}_{CE}$ regularizes training. Integrated with the segmentation backbone, we elaborate on AMContrast3D. The Fig.~\ref{fig:framework} consists of an ambiguity estimation framework (AEF) extracting $\mathcal{P}^s$ at encoding stages to generate ambiguity $\mathcal{A}^s = \{a_i\}_{i=1}^{n^s}$. Based on AEF, an ambiguity-aware margin generator (MG) calculates margin $\mathcal{M}^s = \{m_i\}_{i=1}^{n^s}$, where $a_i, m_i \in \mathbb{R}^1$. We regularize $\mathcal{F}^s$ at decoding stages by an adaptive margin contrastive objective $\mathcal{L}_{AM}^s$, where this objective constructs discriminated embeddings for segmentation.

\subsection{Ambiguity Estimation Framework (AEF)}
\label{sec:method_AAA_AEF}

\begin{figure}[!t]
\centering
\includegraphics[width=3.4in]{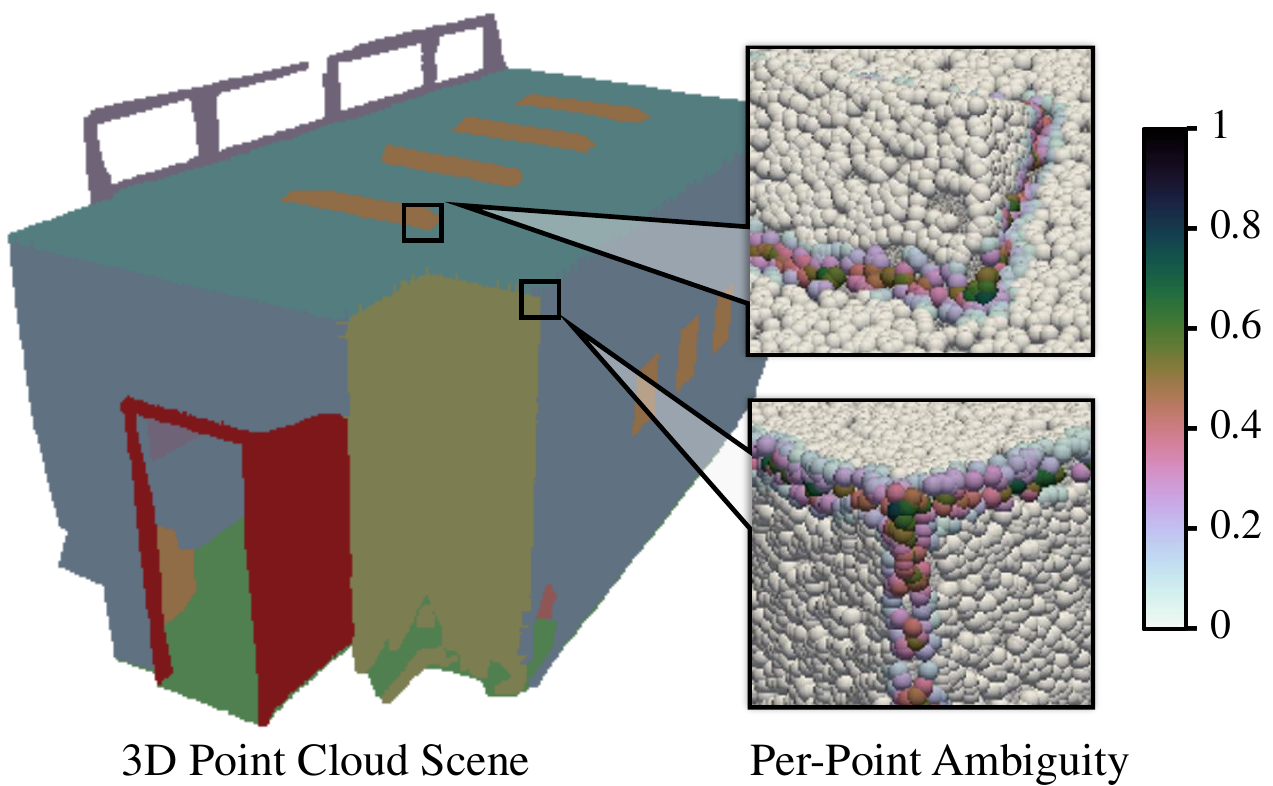}
\caption{Ambiguity visualization. Different semantic classes are shown on 3D point cloud scene, where the per-point ambiguity is colored by a map in $[0,1]$.}
\label{fig:ambiguity}
\end{figure}

A point in 3D space is typically associated with neighboring geometric information, and we propose a framework to extract such information. Given an $i^{th}$ point, we compute Euclidean distances to define its $K$-nearest neighbor points in a set $\mathcal{N}_i^+$. Within $\mathcal{N}_i^+$, most $j^{th}$ neighbor points are intra-points with the same semantic label as $y_j = y_i$, resulting in unambiguous embedding learning. If some $k^{th}$ neighbor points are inter-points with $y_k \neq y_i$, we reallocate them to a new set as $\mathcal{N}_i^-$, which means the $i^{th}$ point is in a transition region, encountering both a positive impact from the intra-class and a negative impact from the inter-class. Intuitively, under a fixed neighboring  size as $K = |\mathcal{N}_i^+| + |\mathcal{N}_i^-|$, larger $|\mathcal{N}_i^-|$ negatively reflects higher ambiguity. Inspired by the closeness centrality in graphs~\cite{veremyev2019finding},~\cite{matsypura2023finding},~\cite{nasirian2020detecting}, which measures the average inverse distance of a node to surrounding nodes, we treat each point as a node in an unconnected graph and design two kinds of closeness centrality by position embeddings $p_i, p_j, p_k$ as:
\begin{eqnarray}
\label{eq:cc}
\begin{aligned}
cc_i^+ = (\frac{\sum_{j=1}^{|\mathcal{N}_i^+|}(p_i - p_j)^2}{|\mathcal{N}_i^+|})^{-1} = \frac{|\mathcal{N}_i^+|}{d_i^+}, \\
cc_i^- = (\frac{\sum_{k=1}^{|\mathcal{N}_i^-|}(p_i - p_k)^2}{|\mathcal{N}_i^-|})^{-1} = \frac{|\mathcal{N}_i^-|}{d_i^-} ,
\end{aligned}
\end{eqnarray}
where the $i^{th}$ point has different compactness with all intra-points as $cc_i^+$ and with all inter-points as $cc_i^-$. Closeness centrality highly correlates with point importance to reflect its compact relation within a neighborhood. Significantly discrepant $cc_i^+$ and $cc_i^-$ are derived from various point numbers and irregular position embeddings of intra-points and inter-points. We find that such a discrepancy of a point can be formulated as a paired subtraction between $cc_i^+$ and $cc_i^-$ in a local neighborhood, which indicates a rational proxy for the ambiguous level. Therefore, we leverage $cc_i^+ - cc_i^-$ into a monotonic decreasing curve, which is formulated as an inverse sigmoid function $\mathcal{G}(cc_i^+, cc_i^-) \in (0,1)$:
\begin{eqnarray}
\label{eq:G}
\mathcal{G}(cc_i^+, cc_i^-) = \frac{1}{1+exp(\beta \cdot (cc_i^+-cc_i^-))},
\end{eqnarray}
where $\beta$ is a tuning parameter. Concretely, a large $cc_i^+$ and a small $cc_i^-$ present on a point with low ambiguity approaching $0$; on the contrary, high ambiguity approaches $1$. The minimum $|\mathcal{N}_i^+|$ is $1$, which is a possible circumstance meaning that a neighborhood only contains $1$ intra-point as the $i^{th}$ point itself, and the other points are inter-points. We define such a point as extremely ambiguous, with the maximum value equaling to 1. To consider all circumstances, a piece-wise function estimates ambiguity $a_i \in \mathcal{A}^s$ at each stage as:
\begin{eqnarray}
\label{eq:A}
a_i = 
\begin{cases}
0 & \text{if } |\mathcal{N}_i^+| = K, \\
\mathcal{G}(cc_i^+, cc_i^-) & \text{if } 1 < |\mathcal{N}_i^+| < K, \\
1 & \text{if } |\mathcal{N}_i^+| = 1, 
\end{cases}
\end{eqnarray}
where Fig.~\ref{fig:ambiguity} visualizes the ambiguity $a_i$ in a point cloud scene. Low-ambiguity, semi-ambiguity, and high-ambiguity points with $a_i \in (0,1]$ are surrounded by various numbers of inter-points in transition regions. We focus on these points to assign point-level contrast dynamically to stabilize the training.

\subsection{Adaptive Margin Contrastive Learning}
\label{sec:method_AAA_CL}

\begin{algorithm}[!t]
    \caption{AMContrast3D of the $i^{th}$ point at stage $s$.} \label{code:3AM}
    \begin{algorithmic}[1]%
    \REQUIRE  $p_i \in \mathcal{P}^s$, $f_i \in \mathcal{F}^s$, $y_i \in \mathcal{Y}^s$, size $K$, temperature $\tau$
    \ENSURE $a_i \in \mathcal{A}^s, m_i \in \mathcal{M}^s, emb_{ij}, emb_{ik}$
    \STATE Neighbor points $\leftarrow p_i$ \COMMENT{\texttt{Position space} $\mathcal{P}^s$}
    \FOR{$j,k$ in $K$}
        \STATE Compute $|\mathcal{N}_i^+|$, $d_i^+$ \COMMENT{\texttt{Intra:} $y_i = y_j$}
        \STATE Compute  $|\mathcal{N}_i^-|$, $d_i^-$ \COMMENT{\texttt{Inter:} $y_i \neq y_k$}   
    \ENDFOR
    \IF{$\mathcal{N}_i^+,\mathcal{N}_i^- \neq \emptyset$}
        \STATE Generate $cc_i^+$ and $cc_i^-$
    \ENDIF
    \STATE Update $\mathcal{A}^s$ from estimated ambiguity $a_i$
    \STATE Update $\mathcal{M}^s$ from margin $m_i$ with $a_i$ awareness
    \STATE $(f_i, f_j), (f_i, f_k)  \leftarrow f_i, f_j, f_k$  \COMMENT{\texttt{Feature space} $\mathcal{F}^s$}
    \FOR{$j,k$ in $K$}
        \STATE $emb_{ij} \leftarrow sim(f_i, f_j), \tau, m_i$ \COMMENT{\texttt{Intra:} $y_i = y_j$}
        \STATE $emb_{ik} \leftarrow sim(f_i, f_k), \tau$ \COMMENT{\texttt{Inter:} $y_i \neq y_k$} 
    \ENDFOR
    \end{algorithmic}
\end{algorithm}

\subsubsection{Revisiting Contrastive Learning} Following the setup of a temperature parameter $\tau$ controlling the contrast~\cite{ oord2018representation},~\cite{khosla2020supervised}, contrastive learning encourages intra-class compactness and inter-class separability. A supervised contrastive objective for the $i^{th}$ point is formulated as:
\begin{eqnarray}\label{eq:LOSS_CL}
    - log\frac{\sum_{j=1}^{|\mathcal{N}_i^+|}exp(\frac{sim(f_i, f_j)}{\tau})}{\sum_{j=1}^{|\mathcal{N}_i^+|}exp(\frac{sim(f_i, f_j)}{\tau}) + \sum_{k=1}^{|\mathcal{N}_i^-|}exp(\frac{sim(f_i, f_k)}{\tau})},
\end{eqnarray} 
where the objective intensifies the discrimination on feature similarities, maximizing intra-pair $sim(f_i, f_j)$ while minimizing inter-pair $sim(f_i, f_k)$. The~(\ref{eq:LOSS_CL}) shares a common ground with the cross-entropy objective~\cite{he2020momentum}, and the decision margins can modify the cross-entropy objective~\cite{deng2019arcface},~\cite{wang2018cosface},~\cite{li2022towards}. Thus, the decision boundaries $DB^+$ for intra-pairs and $DB^-$ for inter-pairs in~(\ref{eq:LOSS_CL}) are defined as:
\begin{eqnarray}
\label{eq:DB_0}
\begin{aligned}
DB^+: sim(f_i, f_j) - sim(f_i, f_k) \ge 0, \\
DB^-: sim(f_i, f_j) - sim(f_i, f_k) \le 0.
\end{aligned}
\end{eqnarray}
We find that~(\ref{eq:LOSS_CL}) and~(\ref{eq:DB_0}) pose two limitations: 1) The margin is $0$, which means $DB^+$ and $DB^-$ are adjacent without discrimination. 2) The objective only targets feature embeddings in $\mathcal{F}^s$ but disregards position embeddings in $\mathcal{P}^s$.

\subsubsection{Ambiguity-aware Margin Generator (MG)} We address these limitations by margins. Intuitively, a fixed positive margin directly generates the expansion between $DB^+$ and $DB^-$, forcing all points to reach a complicated contrastive objective identically. Since individual points with various ambiguities require adaptive training objectives, for the generator to make use of the ambiguities, we explicitly inject $a_i \in (0,1]$ as margin confidence to generate adaptive margin $m_i$ as:
\begin{eqnarray}
\label{eq:AMG_mi}
m_i = \mu \cdot a_i+ \nu,
\end{eqnarray}
where $\mu$ and $\nu$ are the scale and bias parameters, respectively. As shown in Fig.~\ref{fig:teaser}, we control feature discrimination between intra-class and inter-class through point-level margins by the ambiguity-aware generator. Margins involve positive, zero, and negative values in a principled manner: positive margins with penalized separations between $DB^+$ and $DB^-$ heighten objectives for low-ambiguity points; zero margins remain for semi-ambiguity points; negative margins allow moderate feature mixtures with easy objectives, properly degrading the training difficulty for high-ambiguity points. Concretely, we quantify the degree of feature similarity using cosine similarity, which is the dot product of the feature vectors divided by the product of lengths, $i.e.,$ $sim(f_i,f_j)= \frac{f_i \cdot f_j}{\left \| f_i \right \| \left \| f_j \right \|} \in [-1,1]$ is the intra-pair similarity. As a discriminative hyperplane, margin $m_i$ dynamically shifts decision boundaries as:
\begin{eqnarray}
\label{eq:DB_mi}
\begin{aligned}
DB^+: sim(f_i,f_j) - sim(f_i,f_k) \ge m_i, \\
DB^-: sim(f_i,f_j) - sim(f_i,f_k) \le m_i.
\end{aligned}
\end{eqnarray}
Fig.~\ref{fig:framework} illustrates a red dotted line from ambiguities $a_i \in \mathcal{A}^s$ to margins $m_i \in \mathcal{M}^s$, connecting position embeddings in $\mathcal{P}^s$ and feature embeddings in $\mathcal{F}^s$. As a result, margins modify the contrastive objective to provide adaptive training difficulty for each point, stabilizing the overall training.

\subsubsection{Contrastive Optimization}

We optimize the contrastive learning by encouraging the intra-similarity $sim(f_i, f_j)$ to be larger than the inter-similarity $sim(f_i, f_k)$ plus the margin $m_i$. To satisfy~(\ref{eq:DB_mi}), we generalize the contrastive embeddings $emb_{ij}$ for intra-pairs and $emb_{ik}$ for inter-pairs as:
\begin{eqnarray}
\label{eq:EMB_mi}
\begin{aligned}
    emb_{ij} = exp(\frac{sim(f_i, f_j) - m_i}{\tau}), \\
    emb_{ik} = exp(\frac{sim(f_i, f_k)}{\tau}),
\end{aligned}
\end{eqnarray}
where $\tau$ controls the contrast. Algorithm~\ref{code:3AM} explains $emb_{ij}$ and $emb_{ik}$, from which we develop an adaptive margin contrastive objective as $\mathcal{L}_{AM}^s$. Suppose $n^s$ is the point number of low-$a_i$, semi-$a_i$, and high-$a_i$ at stage $s$, the $\mathcal{L}_{AM}^s$ is formulated as:
\begin{eqnarray}
\label{eq:LOSS_DMCL}
    \mathcal{L}_{AM}^s = \frac{1}{n^s}\sum_{i=1}^{n^s} - log\frac{\sum_{j=1}^{|\mathcal{N}_i^+|}emb_{ij}}{\sum_{j=1}^{|\mathcal{N}_i^+|}emb_{ij} + \sum_{k=1}^{|\mathcal{N}_i^-|}emb_{ik}}.
\end{eqnarray}
Our joint training uses $\mathcal{L}_{AM}^s$ and the cross-entropy objective $\mathcal{L}_{CE}$, where $\mathcal{L}_{CE}$ regularizes the label prediction $\tilde{y}_i$ based on the ground-truth $y_i$, and the segmentation objective $\mathcal{L}_{SEG}$ is:
\begin{eqnarray}
\label{eq:LOSS-T}
\mathcal{L}_{SEG} = \lambda \cdot \mathcal{L}_{CE} + (1-\lambda) \cdot\sum_{s}\mathcal{L}_{AM}^s,
\end{eqnarray}
with a balanced parameter $\lambda$. The objective $\mathcal{L}_{AM}^s$ maximizes $emb_{ij}$ and minimizes $emb_{ik}$, resulting in an anchor point to be similar to its intra-points compared to its inter-points.

\section{AMContrast3D++}
\label{sec:method_MMM}

\subsection{Overall Pipeline}

\begin{figure}[!t]
\centering
\includegraphics[width=3.4in]{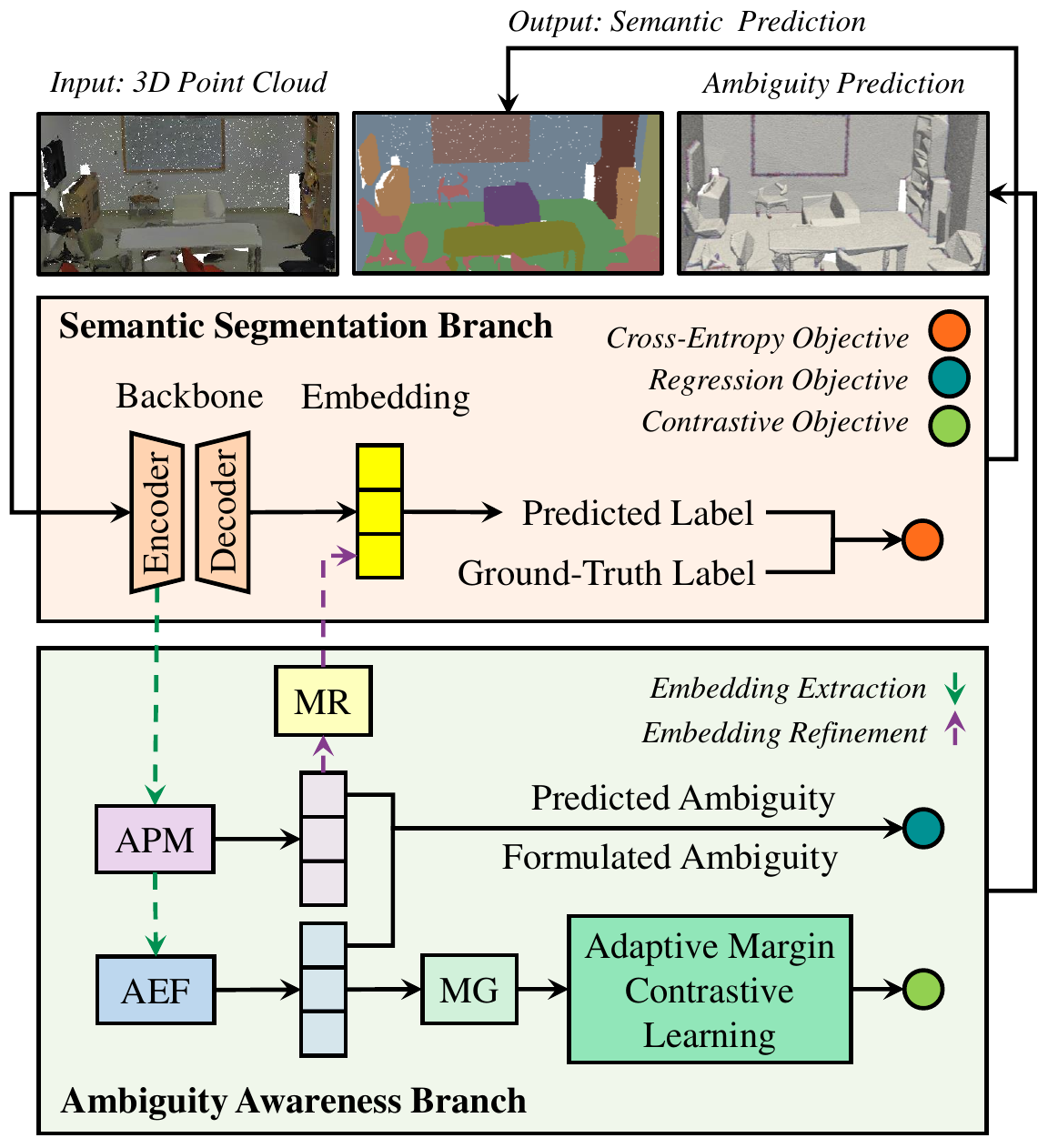}
\caption{Overall pipeline of AMContrast3D++ with two branches and three objectives. The semantic segmentation branch receives a 3D point cloud for semantic prediction. The ambiguity awareness branch consists of an ambiguity estimation framework (AEF), an ambiguity prediction module (APM), a margin generator (MG), adaptive margin contrastive learning, and a masked refinement (MR) mechanism, from which the segmentation is benefited from the ambiguity awareness branch under joint training.}
\label{fig:overview}
\end{figure}

We design a two-branch framework with three joint training objectives, as illustrated in Fig.~\ref{fig:overview}. The primary semantic segmentation branch ingests 3D point clouds to generate semantic predictions. Building on this foundation, we introduce an auxiliary ambiguity awareness branch designed to predict point-level ambiguities. Compared with existing methods, our auxiliary branch facilitates the primary branch, benefiting the segmentation by incorporating rich ambiguity awareness into the embedding learning process. In the ambiguity awareness branch, we first design a simple yet effective module, named the ambiguity prediction module (APM), which ingests point embeddings $\mathcal{P}^s$ and $\mathcal{F}^s$ at encoding stages. APM outputs predicted ambiguity $\tilde{\mathcal{A}}^s = \{\tilde{a}_i\}_{i=1}^{n^s}$ supervised by $\mathcal{A}^s$ through a regression objective $\mathcal{L}_{REG}^s$, where $\tilde{a}_i \in \mathbb{R}^1$. We further leverage $\tilde{a}_i$ to construct per-point masks and propose a masked refinement (MR) mechanism to refine embeddings $\mathcal{F}^s$ at decoding stages. The following subsections provide the details.

\subsection{Ambiguity Prediction Module (APM)}
\label{sec:method_MMM_APM}

\begin{figure*}[!t]
\centering
\includegraphics[width=7.0in]{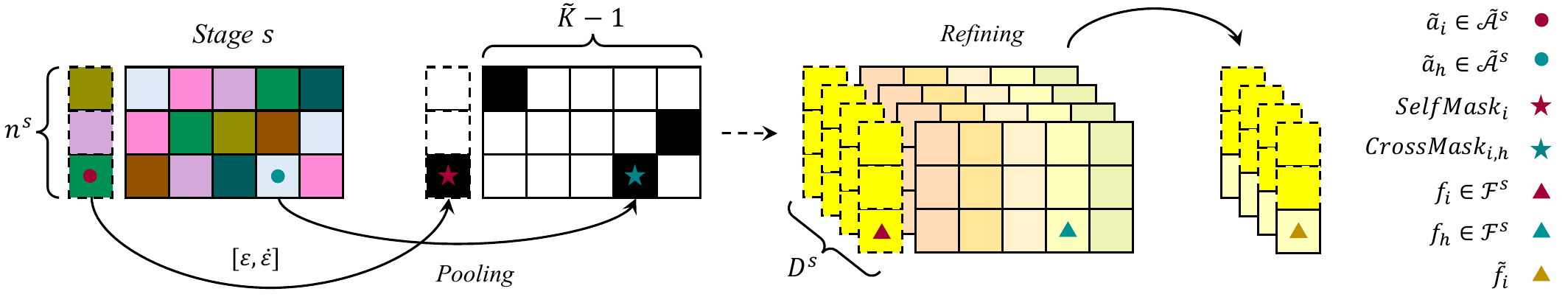}
\caption{Masked refinement. The ambiguity $\widetilde{a}_i \in \widetilde{\mathcal{A}}^s$ is generated by APM for each of the $n^s$ points at stage $s$. We leverage the ambiguity to refine the $D^s$-dim feature embedding $f_i \in \mathcal{F}^s$. We first design per-point $SelfMask_i$ based on the predicted ambiguity $\widetilde{a}_i$ with thresholds $\varepsilon$ and $\dot{\varepsilon}$, where $\widetilde{a}_i \in [\varepsilon, \dot{\varepsilon}]$ means a high ambiguity for the $i^{th}$ point, and we note it with '\textcolor{purple}{$\bigstar$}', indicating the necessity of ambiguity awareness. Next, we conduct retrieval of the nearest neighboring points for the $i^{th}$ point to group its nearest point ambiguity $\widetilde{a}_h$ and its nearest feature embedding $f_h$, where $h=1,2,...,\widetilde{K}-1$. Through the pooling operation, we define $CrossMask_{i,h}$ with '\textcolor{teal}{$\bigstar$}' to locate a neighboring point with the lowest ambiguity, and update relevant embeddings $\widetilde{f}_i$ during the refining process.}
\label{fig:refine}
\end{figure*}

\subsubsection{Regression Architecture}

APM addresses the prediction of point ambiguity as a regression task. Upon encoding stage $s$, we denote each corresponding block as $Block^s$ in APM. $Block^s$ comprises a sequence of $T$ layers, and each layer involves operations including vector concatenation $\odot$, multi-layer perceptron $MLP_{{\theta}^s}$, batch normalization $Norm(\cdot)$, and sigmoid activation $\sigma(\cdot)$. It first ingests $3$-dim $p_i$ and $D^s$-dim $f_i$ as input, where position embedding $p_i \in \mathcal{P}^s$ and feature embedding $f_i \in \mathcal{F}^s$ provide point properties to enhance geometric knowledge. The concatenated input vector is:
\begin{eqnarray}
\label{eq:APM_input}
z_i = p_i \odot f_i, \text{ } z_i \in \mathbb{R}^{3+D^s},
\end{eqnarray}
where the input dimension of $Block^s$ is in alignment with each varied feature dimension $D^s$ at stage $s$. The input vector passes through connected layers, and we construct the overall architecture to output the ambiguity prediction $\widetilde{a}_i$ as:
\begin{eqnarray}
\label{eq:APM_mlp}
\widetilde{a}_i = \sigma (Norm(MLP_{{\theta}^s} (z_i))), \text{ } \theta^s=\{W^{(t)},b^{(t)}\}_{t=1}^{T},
\end{eqnarray}
where total $T$ connected layers are parameterized by learnable ${\theta}^s$ including weight matrix $W^{(t)}$ and bias vector $b^{(t)}$ for $t = 1, 2, ..., T$. Specifically, the ambiguity transformation of $\widetilde{a}_i^{(t)}$ for layer $t$ is expressed as:
\begin{eqnarray}
\label{eq:APM_layer}
\widetilde{a}_i^{(t)} = \sigma (Norm(W^{(t)} \cdot \widetilde{a}_i^{(t-1)} + b^{(t)})).
\end{eqnarray}
Accordingly, $Block^s$ culminates with a $1$-dim prediction:
\begin{eqnarray}
\label{eq:APM_output}
\widetilde{a}_i = \widetilde{a}_i^{(T)}, \text{ } \widetilde{a}_i \in \mathbb{R}^{1},
\end{eqnarray}
where $Block^s(\mathcal{P}^s, \mathcal{F}^s) \rightarrow \widetilde{\mathcal{A}}^s$ with $\widetilde{a}_i \in \widetilde{\mathcal{A}}^s$ is indicative of predicted ambiguities for each stage $s$, providing essential information during training. Meanwhile, it is lightweight involving marginal computation costs.

\subsubsection{Regression Objective}

We develop a regression objective $\mathcal{L}_{REG}^s$ to regularize the training of $Block^s$. This objective is defined as the mean absolute error to compute the discrepancy between the predicted ambiguity $\widetilde{a}_i \in \widetilde{\mathcal{A}}^s$ and formulated ambiguity $a_i \in \mathcal{A}^s$ for each $i^{th}$ point:
\begin{equation}
\label{eq:LOSS-REG}
\mathcal{L}_{REG}^s = \frac{1}{n^s} \sum_{i=1}^{n^s} \left\| \widetilde{a}_i - a_i\right\|, \\
\end{equation}
where stage $s$ has $n^s$ points. Both $\widetilde{a}_i$ and $a_i$ are ranging from $0$ to $1$, and APM predicts $\widetilde{\mathcal{A}}^s$ with supervisory signals of $\mathcal{A}^s$ from AEF in~(\ref{eq:A}), approaching to desired ambiguities. Different from AEF which abstracts local geometries by physical calculation, APM explores block extraction. AEF determines point-level training difficulty, while APM indicates confidence in refining ambiguous embeddings during training, and APM infers point ambiguities after completing training, thereby concentrating more informative cues in spatial regions.

\subsection{Masked Refinement (MR)}
\label{sec:method_MMM_MR}

MR is a mechanism following APM to fuse predicted ambiguities with learnable embeddings, enabling the segmentation branch to enhance semantics for feature discrimination. Fig.~\ref{fig:refine} illustrates details of MR, which depicts refining based on two ambiguity masks: $SelfMask_i$ for anchor points through pre-defined thresholds and $CrossMask_i$ for neighboring points through pooling operations.

\subsubsection{Ambiguity Mask for Anchor Points} 
To distinguish the ambiguity levels for local points, we first propose an ambiguity mask $SelfMask_i \in \mathbb{R}^1$ for the $i^{th}$ point, which involves self characteristics, and the mask is formulated as:
\begin{eqnarray}
\label{eq:SelfMask}
SelfMask_i =
\begin{cases}
1 & \text{if } \widetilde{a}_i \in [\varepsilon, \dot{\varepsilon}], \\
0 & \text{if } \widetilde{a}_i \in [0, \varepsilon) \cup (\dot{\varepsilon},1],
\end{cases}
\end{eqnarray}
where $\varepsilon$ and $\dot{\varepsilon}$ are thresholds within a range of $[0,1]$ to qualify the ambiguity level. We define $\varepsilon \le \dot{\varepsilon}$, and a point with ambiguity $\widetilde{a}_i \in [\varepsilon, \dot{\varepsilon}]$ reflects high confidence of embedding refinement, where the point has a mask as $1$.

\subsubsection{Ambiguity Mask for Neighboring Points} 
We conduct mining for $i^{th}$ anchor point with a neighboring size as $\widetilde{K}$, so it has $\widetilde{K}-1$ nearest neighboring points. We index each neighboring point as $h$ with related ambiguity as $\widetilde{a}_h$. Therefore, each anchor point has a set of neighboring ambiguities:
\begin{eqnarray}
\label{eq:cross_set}
\{\widetilde{a}_h\}_{h=1}^{\widetilde{K}-1}, \text{ } h = 1,2,...,\widetilde{K}-1.
\end{eqnarray}
Besides, it is important to note that the ambiguity of the anchor point is $\widetilde{a}_i$, and we further conduct minimum pooling on the set to extract the lowest ambiguity level defined as $\widetilde{a}_i^{\ast}$ from all neighboring ambiguities:
\begin{eqnarray}
\label{eq:cross_spooling}
\widetilde{a}_i^{\ast} = Pooling(\{\widetilde{a}_h\}_{h=1}^{\widetilde{K}-1}).
\end{eqnarray}
Therefore, each $h^{th}$ neighboring point has an ambiguity mask $CrossMask_{i,h}$ based on $\widetilde{a}_i^{\ast}$, from which we conduct stepwise arrangement to assign masks as $0$ or $1$. Specifically, $1$ indicates a neighboring point satisfying the lowest ambiguity level:
\begin{eqnarray}
\label{eq:CrossMask}
\begin{aligned} 
CrossMask_{i,h} =
\begin{cases}
1 & \text{if } \widetilde{a}_h =\widetilde{a}_i^{\ast}, \\
0 & \text{otherwise},
\end{cases}
\end{aligned}
\end{eqnarray}
where the set of neighboring masks $CrossMask_i \in \mathbb{R}^{\widetilde{K}-1}$ for each $i^{th}$ point is represented as:
\begin{eqnarray}
\label{eq:CrossMask_set}
\begin{aligned} 
CrossMask_i = \{CrossMask_{i,h}\}_{h=1}^{\widetilde{K}-1}.
\end{aligned}
\end{eqnarray}
Based on the ambiguous situations of $SelfMask_i$ for anchor points and $CrossMask_i$ for neighboring points, we further conduct a refining scheme on learnable embeddings. 

\subsubsection{Refining Scheme}

Each $i^{th}$ anchor point has a learnable embedding $f_i \in \mathbb{R}^{D^s}$ from stage $s$, where its $h^{th}$ neighboring embedding $f_h$ is leveraged to refine $f_i$. We first calculate the summation of the element-wise vector product as $f_i^{\ast}$:
\begin{eqnarray}
\label{eq:EmbRefine_sum}
\begin{aligned}
f_i^{\ast} = \sum_{h=1}^{\widetilde{K}-1} f_h \cdot CrossMask_{i,h}.
\end{aligned}
\end{eqnarray}
Next, we generate a representative embedding $\widetilde{f}_i$ for each $f_i$, where it is kept the same as $f_i$ or refined as new embedding:
\begin{eqnarray}
\label{eq:EmbRefine_comb}
\begin{aligned}
\widetilde{f}_i = f_i \cdot (1 - SelfMask_i) \ + f_i^{\ast} \cdot SelfMask_i \\
 =
\begin{cases}
f_i^{\ast} & \text{if } SelfMask_{i}=1 \text{ in~(\ref{eq:SelfMask})}, \\
f_i & \text{if } SelfMask_{i}=0 \text{ in~(\ref{eq:SelfMask})}.
\end{cases}
\end{aligned}
\end{eqnarray}
To maintain the original embedding with ambiguity awareness, we conduct partial refinement for the $i^{th}$ anchor point as:
\begin{eqnarray}
\label{eq:EmbRefine_3}
f_i =  \gamma \cdot \widetilde{f}_i + (1 - \gamma) \cdot f_i,
\end{eqnarray}
where a balanced parameter $\gamma \in [0,1]$ represents the refining rate. In general, a large $\gamma$ denotes a high rate of refining, and $\gamma$ approaches $1$ leveraging a full embedding refinement. The value of $\gamma$ around $0$ represents an insignificant refinement.

\subsection{Joint Training Objective}
We jointly train the segmentation backbone in the semantic segmentation branch and the APM in the ambiguity awareness branch in parallel. Defining $\omega$ as the balanced parameter, we construct a joint objective $\mathcal{L}$ for the overall optimization as:
\begin{eqnarray}
\label{eq_loss}
\mathcal{L} = \mathcal{L}_{SEG} + \omega \cdot \sum_{s} \mathcal{L}_{REG}^s,
\end{eqnarray}
where the backbone is regularized by segmentation objective $\mathcal{L}_{SEG}$ in~(\ref{eq:LOSS-T}), including a common cross-entropy objective $\mathcal{L}_{CE}$ at the output stage and our adaptive margin contrastive objective $\mathcal{L}_{AM}^s$ at all stages. APM is regularized by regression objective $\mathcal{L}_{REG}^s$ in~(\ref{eq:LOSS-REG}).

\section{Experiment}
\label{sec:exp}
In this section, we first introduce our experimental setting (Section~\ref{sec:exp_A_setting}). We conduct extensive experimental results to evaluate the baseline, AMContrast3D, and AMContrast3D++ by quantitatively comparing many state-of-the-art methods and qualitatively summarizing key observations (Section~\ref{sec:exp_B_results}). We also thoroughly analyze the design choice and parameter sensitivity in ablation studies (Section~\ref{sec:exp_C_ablation}) and provide additional discussions (Section~\ref{sec:exp_D_discussion}).

\subsection{Experimental Setting}
\label{sec:exp_A_setting}

\subsubsection{Datasets and Evaluation Metrics} We use two indoor scene datasets, S3DIS~\cite{armeni2017joint} and ScanNet~\cite{dai2017scannet}. S3DIS covers $271$ large-scale indoor rooms in $6$ areas with total semantic classes $C$ as $13$. We take Area 5 for inference and others for training. ScanNet contains cluttered indoor scenes annotated with classes $C$ as $20$. It contains $1613$ cluttered scans, which are split into $1201$ training scans, $312$ validation scans, and $100$ test scans. We evaluate our methods on three metrics, OA ($\%$), mACC ($\%$), and mIoU ($\%$). OA is the overall accuracy, mACC is the mean accuracy within each class, and mIoU is the mean intersection over union.

\subsubsection{Model Architectures}

We adopt the same architecture on both datasets. For the segmentation backbone, we use deep MLPs-based PointNeXt~\cite{qian2022pointnext}, which has a stem MLP with channel size as $64$, InvResMLP blocks, and SA blocks from PointNet++~\cite{qi2017pointnet++}. The initial feature dimension $D$ is $4$. We have encoding-decoding stages indexed by $s=1,2,3,4$ with the feature dimension $D^s$ as $(64, 128, 256, 512)$. Based on the label mining strategy from~\cite{tang2022contrastive}, we integrate label embeddings at each stage. For the APM regression architecture $Block^s$ with $T$ layers, we set the layer number $T$ as $6$ and $(3+D^s, 32, 16, 8, 4, 2, 1)$ as the channel dimension. 

\subsubsection{Implementation Details}

We use an initial learning rate of $0.01$ with $150$ epochs for a training episode. All model variants are implemented with the PyTorch toolkit and two $24$GB GPUs. The balanced parameters $\lambda$ and $\omega$ for training objectives are $0.1$ and $0.01$, respectively. The neighboring size $K$ is $24$ and the parameter $\beta$ is $0.04$. The thresholds $\varepsilon$ and $\dot{\varepsilon}$ are $0.9$ and $1$. For S3DIS dataset, our model is trained using a batchsize of $5$ per GPU, and input points are downsampled with $n$ as $24000$ per batch. The margin $m_i$ adjusts for each point with scale parameter $\mu$ as $-1$ and bias parameter $\nu$ as $0.5$. Thus, points with low $a_i \in (0,0.5)$ lead to $m_i>0$ for expanded decision boundaries; points with $a_i = 0.5$ have $m_i=0$ for adjacent decision boundaries; points with high $a_i \in (0.5,1]$ lead to $m_i<0$ with mixed decision boundaries. The temperature $\tau$ is $0.3$, and rate $\gamma$ is $1$. The size $\widetilde{K}$ is $12$. For ScanNet dataset, the batchsize is $2$ per GPU, and the point number $n$ is $64000$ per batch. The settings of $\tau$, $\gamma$, and $\widetilde{K}$ are $0.5$, $0.6$, and $8$, respectively. Margin $m_i$ has $\mu$ as $-1$ and $\nu$ as $0.6$. Thereby, points with $a_i \in (0,0.6)$, $a_i = 0.6$, and $a_i \in (0.6,1]$ lead to positive, zero, and negative $m_i$.

\begin{table*}[t]
\begin{center}
\caption{3D semantic segmentation results of each semantic class on the S3DIS dataset (Area 5). We report the per-class IoU ($\%$) for $13$ semantic classes. The best and the second-best results are \textbf{bolded} and \underline{underlined}, respectively. We thoroughly compare our methods with the baseline by indicating the performance improvements (up arrow) or declines (down arrow). \label{tab:s3dis_class}}
\begin{tabular}{l|c|c|c|c|c|c|c|c|c|c|c|c|c}
  \toprule
  \makecell[c]{Method} & \textit{ceiling} & \textit{floor} & \textit{wall} & \textit{beam} & \textit{column} & \textit{window} & \textit{door} & \textit{table} & \textit{chair} & \textit{sofa} & \textit{bookcase} & \textit{board} & \textit{clutter}
  \\
  \midrule
  PointNet~\cite{qi2017pointnet}  & 88.8 & 97.3 & 69.8 & \textbf{0.1} & 3.9 & 46.3 & 10.8 & 59.0 & 52.6 & 5.9 & 40.3 & 26.4 & 33.2\\
  PCT~\cite{guo2021pct}  & 92.5 & 98.4 & 80.6 & 0.0 & 19.4 & \textbf{61.6} & 48.0 & 76.6 & 85.2 & 46.2 & 67.7 & 67.9 & 52.3\\
  KPConv~\cite{thomas2019kpconv}   & 92.8 & 97.3 & 82.4 & 0.0 & 23.9 & 58.0 & 69.0 & 81.5 & 91.0 & 75.4 & 75.3 & 66.7 & 58.9 \\
  JSENet~\cite{hu2020jsenet}   & 93.8 & 97.0 & 83.0 & 0.0 & 23.2 & \underline{61.3} & 71.6 & \underline{89.9} & 79.8 & \underline{75.6} & 72.3 & 72.7  &60.4 \\
  CBL~\cite{tang2022contrastive}   & 93.9 & 98.4 & 84.2 & 0.0 & \textbf{37.0} & 57.7 &  71.9 & \textbf{91.7} & 81.8 & \underline{77.8} & 75.6 & 69.1 & 62.9 \\
  \midrule
  \rowcolor{blue!4} PointNeXt~\cite{qian2022pointnext}  & \underline{94.0} & \underline{98.5} & 84.4 & 0.0 & \underline{36.8} & 59.2 & \underline{73.3} & 83.9 & 90.5 & 74.3 & 76.0 & 79.7 & 59.9 \\
  \midrule
  \rowcolor{blue!4} AMContrast3D~\cite{chen2024adaptive}   & \makecell[c]{93.5 \\ \scriptsize \textcolor{gray}{$\downarrow$ 0.5}} & \makecell[c]{\textbf{98.7} \\ \scriptsize \textcolor{teal}{$\uparrow$ 0.2}} & \makecell[c]{\underline{85.1} \\ \scriptsize \textcolor{teal}{$\uparrow$ 0.7}} & 0.0 & \makecell[c]{36.2 \\ \scriptsize \textcolor{gray}{$\downarrow$ 0.6}} & \makecell[c]{58.9 \\ \scriptsize \textcolor{gray}{$\downarrow$ 0.3}} & \makecell[c]{\textbf{79.2} \\ \scriptsize \textcolor{teal}{$\uparrow$ 5.9}} & \makecell[c]{84.0 \\ \scriptsize \textcolor{teal}{$\uparrow$ 0.1}} & \makecell[c]{\underline{92.4} \\ \scriptsize \textcolor{teal}{$\uparrow$ 1.9}} & \makecell[c]{\underline{77.8} \\ \scriptsize \textcolor{teal}{$\uparrow$ 3.5}} & \makecell[c]{\underline{78.6} \\ \scriptsize \textcolor{teal}{$\uparrow$ 2.6}} & \makecell[c]{\textbf{84.9} \\ \scriptsize \textcolor{teal}{$\uparrow$ 5.2}} & \makecell[c]{\underline{63.4} \\ \scriptsize \textcolor{teal}{$\uparrow$ 3.5}} \\
   \midrule
  \rowcolor{blue!4} AMContrast3D++   & \makecell[c]{\textbf{95.0} \\ \scriptsize \textcolor{teal}{$\uparrow$ 1.0}}  & \makecell[c]{\textbf{98.7}\\ \scriptsize \textcolor{teal}{$\uparrow$ 0.2}} & \makecell[c]{\textbf{85.3}\\ \scriptsize \textcolor{teal}{$\uparrow$ 0.9}} & 0.0 & \makecell[c]{36.0 \\ \scriptsize \textcolor{gray}{$\downarrow$ 0.8}} & \makecell[c]{59.0 \\ \scriptsize \textcolor{gray}{$\downarrow$ 0.2}} & \makecell[c]{72.5 \\ \scriptsize \textcolor{gray}{$\downarrow$ 0.8}} & \makecell[c]{83.7 \\ \scriptsize \textcolor{gray}{$\downarrow$ 0.2}} & \makecell[c]{\textbf{92.5}\\ \scriptsize \textcolor{teal}{$\uparrow$ 2.0}} & \makecell[c]{\textbf{80.9} \\ \scriptsize \textcolor{teal}{$\uparrow$ 6.6}} & \makecell[c]{\textbf{79.0} \\ \scriptsize \textcolor{teal}{$\uparrow$ 3.0}} & \makecell[c]{\underline{81.9} \\ \scriptsize \textcolor{teal}{$\uparrow$ 2.2}} & \makecell[c]{\textbf{64.4} \\ \scriptsize \textcolor{teal}{$\uparrow$ 4.5}}\\
  \bottomrule
\end{tabular}
\end{center}
\end{table*}

\begin{table}[t]
\begin{center}
\caption{3D semantic segmentation results on S3DIS dataset (Area 5). We report the OA ($\%$), mACC ($\%$) and mIoU ($\%$). \label{tab:s3dis}} 
\begin{tabular}{l|c|c|c|c}
  \toprule
  \makecell[c]{Method} & Publication & OA & mACC  & mIoU 
  \\
  \midrule
  PointNet++~\cite{qi2017pointnet++} & NeurIPS 2017 & 83.0 & - & 53.5\\
  SPG~\cite{landrieu2018large} & CVPR 2018 & 85.5 & 73.0 & 62.1\\
  KPConv~\cite{thomas2019kpconv} & ICCV 2019 & - & 72.8 & 67.1\\
  PAConv~\cite{xu2021paconv} & CVPR 2021 & - & 73.0 & 66.6\\
  JSENet~\cite{hu2020jsenet} & ECCV 2020 & - & - & 67.7\\
  3D-PAM~\cite{weng2022context} & TMM 2022 & - & - & 68.4 \\
  CBL~\cite{tang2022contrastive} & CVPR 2022  & 90.6 & 75.2 & 69.4\\
  PointTrans.~\cite{zhao2021point} & ICCV 2021 & 90.8 & 76.5 & 70.4\\
  FastPointTrans.~\cite{park2022fast} & CVPR 2022 & - & 77.3 & 70.1\\
  PointMixer~\cite{choe2022pointmixer} & ECCV 2022 & - & 77.4 & 71.4\\
  HGNet~\cite{yao2023hgnet}  & CVPR 2023 & 90.7 & 76.9 & 70.8\\
  SPoTr~\cite{park2023self} & CVPR 2023 & 90.7 & 76.4 & 70.8\\
  PointMetaBase~\cite{lin2023meta} & CVPR 2023 & 90.8 & - & 71.3\\
  PointDif~\cite{zheng2024point} & CVPR 2024 & - & 77.1 & 70.0 \\
  PointCloudMamba~\cite{zhang2024point} & - & 88.2 & 71.0 & 63.4\\
  \midrule
  \rowcolor{blue!4} PointNeXt~\cite{qian2022pointnext} &  NeurIPS 2022 & \makecell[c]{90.4 \\ \textcolor{teal}{\scriptsize $\pm$0.3}} & \makecell[c]{76.1 \\ \textcolor{teal}{\scriptsize $\pm$0.4}} & \makecell[c]{69.8 \\ \textcolor{teal}{\scriptsize $\pm$0.4}}\\
  \midrule
  \rowcolor{blue!4} AMContrast3D~\cite{chen2024adaptive} & ICME 2024 & \makecell[c]{90.9 \\ \textcolor{teal}{\scriptsize $\pm$0.3}} & \makecell[c]{76.9 \\ \textcolor{teal}{\scriptsize $\pm$0.3}} & \makecell[c]{70.8 \\ \textcolor{teal}{\scriptsize $\pm$0.8}} \\
  \midrule
  \rowcolor{blue!4} AMContrast3D++  & - & \makecell[c]{\textbf{91.1} \\ \textcolor{teal}{\scriptsize $\pm$0.1}} & \makecell[c]{\textbf{77.8} \\ \textcolor{teal}{\scriptsize $\pm$0.2}} & \makecell[c]{\textbf{71.4} \\ \textcolor{teal}{\scriptsize $\pm$0.2}} \\
  \bottomrule
\end{tabular}
\end{center}
\end{table}

\begin{table}[t]
\begin{center}
\caption{3D semantic segmentation results on ScanNet dataset. We report the mIoU ($\%$) for the validation set and the test set.\label{tab:scannet}} 
\begin{tabular}{l|c|c|c}
  \toprule
  \makecell[c]{Method}  & Publication & \makecell[c]{mIoU \\ (Val)} & \makecell[c]{mIoU \\ (Test)}
  \\
  \midrule
  PointNet++~\cite{qi2017pointnet++} & NeurIPS 2017 & 53.3 & 33.9 \\
  KPConv~\cite{thomas2019kpconv} & ICCV 2019 & 69.2 & 68.6 \\
  JSENet~\cite{hu2020jsenet} & ECCV 2020 & -  & 69.9 \\
  3D-PAM~\cite{weng2022context} & TMM 2022 & - & 70.6 \\
  CBL~\cite{tang2022contrastive} & CVPR 2022  & -  & 70.5\\
  PointTrans.~\cite{zhao2021point} & ICCV 2021 & 70.6 & - \\
  FastPointTrans.~\cite{park2022fast} & CVPR 2022 & 72.1 & - \\
  PointMetaBase~\cite{lin2023meta} & CVPR 2023 & \textbf{72.8} & 71.4 \\
  \midrule
  \rowcolor{blue!4} PointNeXt~\cite{qian2022pointnext} & NeurIPS 2022 & \makecell[c]{70.1 \\ \textcolor{teal}{\scriptsize $\pm$0.3}} & \makecell[c]{70.0 \\ \textcolor{teal}{\scriptsize $\pm$0.3}} \\
  \midrule
  \rowcolor{blue!4} AMContrast3D~\cite{chen2024adaptive} & ICME 2024 & \makecell[c]{71.2 \\ \textcolor{teal}{\scriptsize $\pm$0.9}} & \makecell[c]{71.2 \\ \textcolor{teal}{\scriptsize $\pm$1.0}}  \\
  \midrule
  \rowcolor{blue!4} AMContrast3D++ & - & \makecell[c]{71.6 \\ \textcolor{teal}{\scriptsize $\pm$0.2}} & \makecell[c]{\textbf{71.7} \\ \textcolor{teal}{\scriptsize $\pm$0.3}}  \\
  \bottomrule
\end{tabular}
\end{center}
\end{table}

\begin{table*}[t]
\begin{center}
\setlength\tabcolsep{3.5pt} 
\caption{Comparison of the baseline and proposed methods, including network module (AEF, MG, APM, MR), training objective, time complexity (training hours, FLOPs), and space complexity (parameter, memory usage). All results are collected with $24000$ input points for S3DIS and $64000$ input points for ScanNet. We conduct experiments with two $24$GB GPUs, and the GPU memory is fully occupied with $95\sim100\%$ volatile GPU-utilization during training.} \label{tab:compare}
\begin{tabular}{l|c|c|c|c|c|c|c|ll|ll|ll|ll}
  \toprule
  \makecell*[c]{\multirow{3}*{Method}} & \multicolumn{4}{c|}{Network module}  & \multicolumn{3}{c|}{Training objective} &  \multicolumn{4}{c|}{Time complexity} & \multicolumn{4}{c}{Space complexity} \\
  \cmidrule(lr){2-16}
   &  \makecell*[c]{\multirow{2}*{AEF}} & \makecell*[c]{\multirow{2}*{MG}} & \makecell*[c]{\multirow{2}*{APM}} & \makecell*[c]{\multirow{2}*{MR}} & \makecell*[c]{\multirow{2}*{$\mathcal{L}_{CE}$}} & \makecell*[c]{\multirow{2}*{$\mathcal{L}_{AM}^s$}} & \makecell*[c]{\multirow{2}*{$\mathcal{L}_{REG}^s$}} & \multicolumn{2}{l|}{Training hours} & \multicolumn{2}{l|}{FLOPs (G)} & \multicolumn{2}{l|}{Parameter (M)} & \multicolumn{2}{l}{Memory usage (GB)} \\
   &   &  &  &  &  &  &  & S3DIS & ScanNet & S3DIS & ScanNet & S3DIS & ScanNet & S3DIS & ScanNet \\
  \midrule
   PointNeXt~\cite{qian2022pointnext} & \textcolor{black}{\ding{55}} & \textcolor{black}{\ding{55}} & \textcolor{black}{\ding{55}} & \textcolor{black}{\ding{55}} & \textcolor{teal}{\Checkmark} & \textcolor{black}{\ding{55}} & \textcolor{black}{\ding{55}} & 28 & 76 & 135.82 & 364.57 & 41.58 & 41.59 & 2$\times$24 & 2$\times$24\\
  AMContrast3D~\cite{chen2024adaptive} & \textcolor{teal}{\Checkmark} & \textcolor{teal}{\Checkmark} & \textcolor{black}{\ding{55}} & \textcolor{black}{\ding{55}} & \textcolor{teal}{\Checkmark} & \textcolor{teal}{\Checkmark} & \textcolor{black}{\ding{55}} & 43 & 115 & 135.82 & 364.57 & 41.58 & 41.59 & 2$\times$24 & 2$\times$24 \\
   AMContrast3D++ & \textcolor{teal}{\Checkmark} & \textcolor{teal}{\Checkmark} & \textcolor{teal}{\Checkmark} & \textcolor{teal}{\Checkmark} & \textcolor{teal}{\Checkmark} & \textcolor{teal}{\Checkmark} & \textcolor{teal}{\Checkmark}  & 47 & 124 & 136.06 & 365.20 & 41.61 & 41.63 & 2$\times$24 & 2$\times$24 \\
  \bottomrule
\end{tabular} 
\end{center}
\end{table*}

\subsection{Results and Analysis}
\label{sec:exp_B_results}

\subsubsection{Segmentation Results on S3DIS}

We compare our methods quantitatively with reference methods for S3DIS dataset. Note that the baseline, PointNeXt, is originally trained using a batchsize of $8$. Due to the computational resource constraints, we retrain PointNeXt with a batchsize of $5$, and we further train AMContrast3D and AMContrast3D++ using the same batchsize for a fair comparison. For inference, we evaluate a set including $3$ metrics as $\{\text{OA}$, $\text{mACC}$, $\text{mIoU}\}$ at
test time. To obtain more reasonable evaluations, we conduct experiments for each method in $4$ random runs, thereby calculating the error bound in the form of $mean_{\pm std}$ for each metric, where $mean$ is the average result of all runs and $std$ is the standard deviation as a dispersion of all the runs in relation to the $mean$. We expect to achieve a high $mean$ with a low $std$, indicating a better performance with strong robustness.

In $4$ experimental runs, PointNeXt acquires $\{90.5\%$, $76.1\%$, $70.1\%\}$, $\{90.0\%$, $76.0\%$, $69.6\%\}$, $\{90.3\%$, $75.6\%$, $69.5\%\}$, and $\{90.8\%$, $76.5\%$, $70.3\%\}$, which has the highest mIoU as $70.3\%$ and a set $\{90.4_{\pm0.3}\%$, $76.1_{\pm0.4}\%$, $69.8_{\pm0.4}\%\}$ as the average OA, mACC, and mIoU in Table~\ref{tab:s3dis}. AMContrast3D acquires $\{91.1\%$, $77.1\%$, $71.8\%\}$, $\{90.9\%$, $76.9\%$, $70.5\%\}$, $\{90.4\%$, $76.5\%$, $70.0\%\}$, and $\{91.0\%$, $77.1\%$, $71.1\%\}$ with the highest mIoU as $71.8\%$ and an average set as $\{90.9_{\pm0.3}\%$, $76.9_{\pm0.3}\%$, $70.8_{\pm0.8}\%\}$. Compared with PointNeXt, our AMContrast3D boosts the results by $0.3\%$, $0.5\%$, and $1.5\%$ for the highest OA, mACC, and mIoU, respectively. Meanwhile, its highest results surpass many methods with obvious margins. However, we observe that its average mIoU is $1\%$ lower than its highest mIoU and its $std$ ($\pm0.8$) is high, which means AMContrast3D is not robust for the entire runs and all results are not closely tight to the $mean$. Equipped with the ambiguity awareness branch, AMContrast3D++ further achieves significant improvements over AMContrast3D with increased $mean$ and decreased $std$. Specifically, its $4$ random runs are $\{91.2\%$, $77.8\%$, $71.7\%\}$, $\{91.1\%$, $78.0\%$, $71.4\%\}$, $\{90.9\%$, $77.7\%$, $71.2\%\}$, and $\{91.1\%$, $77.7\%$, $71.5\%\}$ and generate an average set as $\{91.1_{\pm0.1}\%$, $77.8_{\pm0.2}\%$, $71.4_{\pm0.2}\%\}$. To begin with, we evaluate average $\{\text{OA}$, $\text{mACC}$, $\text{mIoU}\}$, and AMContrast3D++ achieves the leading performances, obtaining the gains of $\{0.2\%$, $0.9\%$, $0.6\%\}$ and $\{0.7\%$, $1.7\%$, $1.6\%\}$ compared to AMContrast3D and PointNeXt, respectively. This observation revolving high $mean$ demonstrates that our key approaches significantly improve segmentation results. In addition, AMContrast3D++ surpasses AMContrast3D with $std$ reductions in terms of OA ($\pm 0.3 \rightarrow \pm 0.1$), mACC ($\pm 0.3 \rightarrow \pm 0.2$), and mIoU ($\pm 0.8 \rightarrow \pm 0.2$). It also surpasses PointNeXt for OA ($\pm 0.3 \rightarrow \pm 0.1$), mACC ($\pm 0.4 \rightarrow \pm 0.2$), and mIoU ($\pm 0.4 \rightarrow \pm 0.2$). As a result, the low $std$ indicates our methods strengthen the robustness. Meanwhile, our AMContrast3D++ achieves better or very competitive segmentation performances with the existing state-of-the-art methods.

Furthermore, we analyze the segmentation results across different semantic classes in Table~\ref{tab:s3dis_class}. The best and the second-best IoU results are bolded and underlined, respectively. S3DIS dataset has $13$ semantic classes, and our methods achieve $9$ best results including \textit{ceiling}, \textit{floor}, \textit{wall}, \textit{door}, \textit{chair}, \textit{sofa}, \textit{bookcase}, \textit{board}, and \textit{clutter}. We demonstrate performance improvements or declines for our methods compared with the baseline. For example, AMContrast3D boosts PointNeXt with per-class IoU improvements on \textit{door} ($\uparrow 5.9\%$), \textit{bookcase} ($\uparrow 2.6\%$), and \textit{board} ($\uparrow 5.2\%$), and AMContrast3D++ also boosts PointNeXt on classes as \textit{chair} ($\uparrow 2.0\%$), \textit{sofa} ($\uparrow 6.6\%$), and \textit{clutter} ($\uparrow 4.5\%$). Meanwhile, compared with AMContrast3D, AMContrast3D++ has improvements on \textit{ceiling} ($\uparrow 1.5\%$), \textit{sofa} ($\uparrow 3.1\%$), \textit{clutter} ($\uparrow 1.0\%$), and many other classes.

To ascertain the effectiveness, we conduct comparisons for PointNeXt, AMContrast3D, and AMContrast3D++ in terms of time and space complexity in Table~\ref{tab:compare}. AMContrast3D designs modules of AEF and MG, and an objective $\mathcal{L}_{AM}^s$ to regularize each training stage. We observe that AMContrast3D achieves notably better performance than PointNeXt under the same number of parameters ($41.58$M) and FLOPs ($135.82$G). While the fully connected layers in APM contribute to the extra learnable parameters (+$0.03$M), and the module MR and the objective $\mathcal{L}_{REG}^s$ also contribute to a slight increase in FLOPs (+$0.24$G), our AMContrast3D++ achieves better performances. These results demonstrate that our methods effectively improve the segmentation performance with small increments in time and space complexity.

\begin{figure}[!t]
\centering
\includegraphics[width=3.4in]{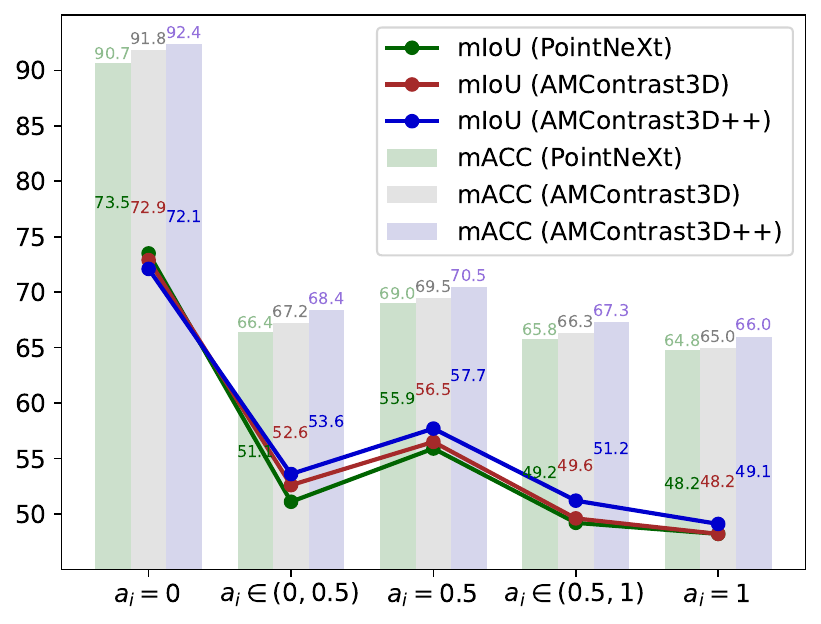}
\caption{Evaluation on each ambiguity level. We report mIoU ($\%$) and mACC ($\%$) results of the baseline (PointNeXt) and our methods (AMContrast3D and AMContrast3D++) on S3DIS with different ambiguity levels, containing $a_i=0$ for unambiguous points, $a_i \in (0,0.5)$ for low-ambiguity points, $a_i=0.5$ for semi-ambiguity points, and $a_i \in (0.5,1) \cup a_i=1$ for high-ambiguity points. AMContrast3D++ achieves leading mIoU (line chart) and mACC (bar chart) results for all ambiguous points with  $a_i \in (0,1]$.}
\label{fig:separate}
\end{figure}

We further perform the evaluation on each ambiguity level, and Fig.~\ref{fig:separate} reports the separated results of mIoU (line chart) and mACC (bar chart). In detail, we choose random runs from PointNeXt, AMContrast3D, and AMContrast3D++ with their overall $\{\text{mIoU}$, $\text{mACC}\}$ as $\{\text{70.3\%}$, $\text{76.5\%}\}$, $\{\text{71.1\%}$, $\text{77.1\%}\}$, $\{\text{71.7\%}$, $\text{77.8\%}\}$, respectively. We assign various ambiguity levels from $0$ to $1$ on individual points, where $a_i=0$ indicates unambiguous points, and these points are generally effortless to segment than ambiguous points. Instead, our methods focus on addressing ambiguous points, which are unclear and need more attention. These ambiguous points are categorized into $a_i \in (0,0.5)$, $a_i=0.5$, $a_i \in (0.5,1)$, and $a_i=1$. We observe that AMContrast3D gains competitive or better results compared to PointNeXt on all levels. Moreover, AMContrast3D++ surpasses AMContrast3D, which achieves leading results by obvious margins, such as the low-ambiguity ($\uparrow 1.0\%$), semi-ambiguity ($\uparrow 1.2\%$), and high-ambiguity ($\uparrow 1.6\%$) on mIoU. AMContrast3D and AMContrast3D++ promote the baseline by $0.6\%$ and $1.3\%$ in terms of the overall mACC.

\begin{figure*}[!t]
\centering
\includegraphics[width=6.6in]{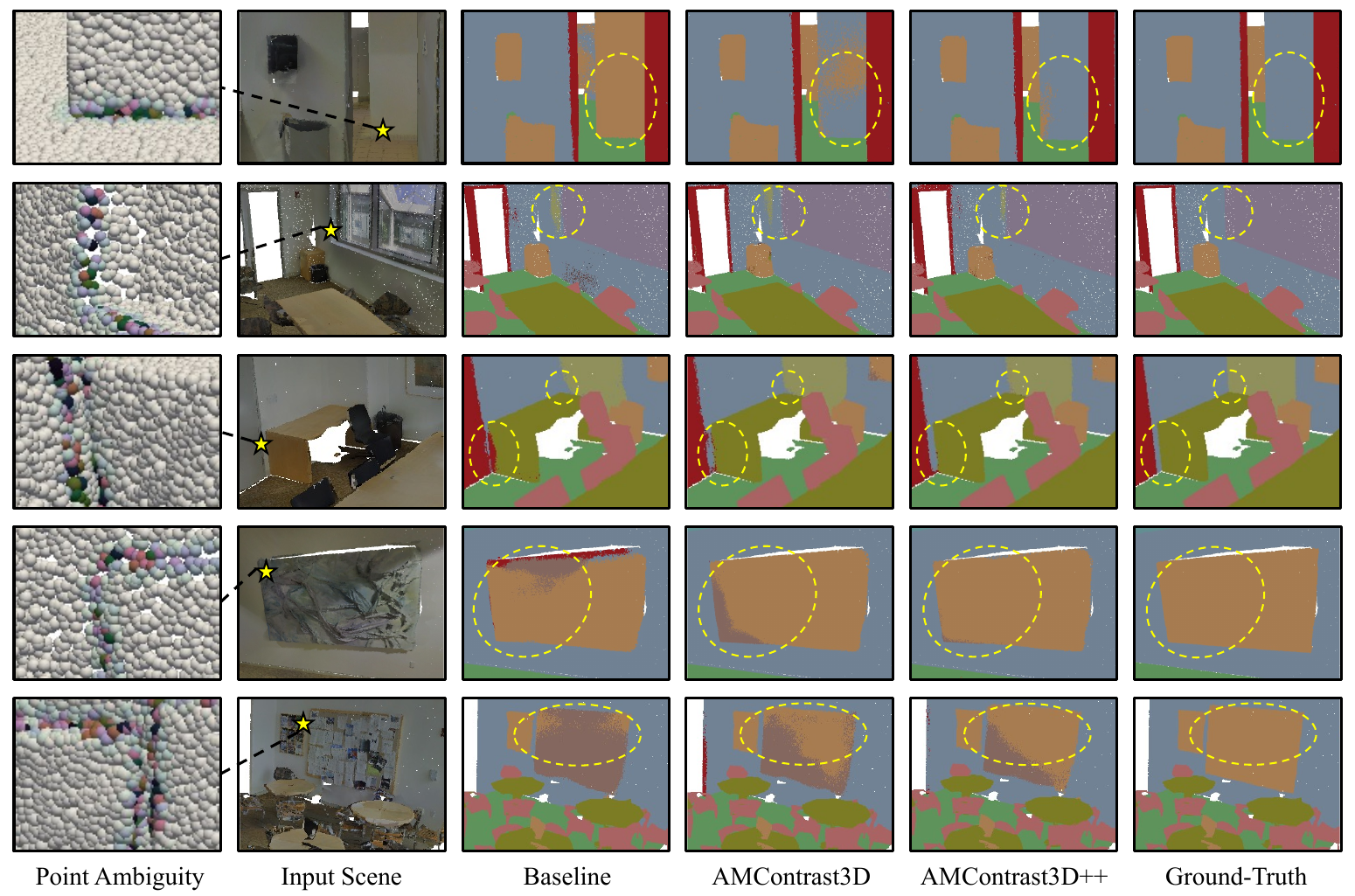}
 \caption{Visualization results on S3DIS dataset (Area 5). We visualize the input scenes, the results predicted by the baseline (PointNeXt),  the results predicted by our method (AMContrast3D and AMContrast3D++), and the ground-truth semantic labels, where the yellow dashed circles indicate the gradual improvements. The yellow stars in the input scenes indicate the representative transition regions connecting different semantic classes, from which we zoom these regions and visualize the detailed ambiguity levels for individual points.}
\label{fig:s3dis}
\end{figure*}

\subsubsection{Segmentation Results on ScanNet}

We fairly compare PointNeXt, AMContrast3D, and AMContrast3D++ using a batchsize of $2$ per GPU with $2$ GPUs for ScanNet. The training of each method is randomly repeated for $4$ times, and we set $\{\text{mIoU (Val)}$, $\text{mIoU (Test)}\}$ for evaluating the validation and test as commonly referenced metrics. We report the error bound through $mean_{\pm std}$ for each metric.

Detailed results of PointNeXt are $\{70.2\%$, $70.2\%\}$, $\{70.3\%$, $70.3\%\}$, $\{69.8\%$, $69.8\%\}$, and $\{70.0\%$, $69.8\%\}$ through $4$ random runs, thereby generating an average as $\{70.1_{\pm0.3}\%$, $70.0_{\pm0.3}\%\}$ in Table~\ref{tab:scannet}. For AMContrast3D, the results are $\{72.5\%$, $72.6\%\}$, $\{71.1\%$, $70.9\%\}$, $\{70.6\%$, $70.5\%\}$, and $\{70.6\%$, $70.8\%\}$, which has an average set as $\{71.2_{\pm0.9}\%$, $71.2_{\pm1.0}\%\}$. We observe that its highest mIoU $72.6\%$ boosts PointNeXt by a large margin as $2.3\%$ and outperforms many recently proposed methods, while its $mean$ as $71.2\%$ only improves PointNeXt by $1.2\%$ with a high $std$ ($\pm1.0$) during the test time. The results of this approach on ScanNet are not robust, demonstrating an alignment with our observations on S3DIS. Our new approach, AMContrast3D++, acquires results $\{71.6\%$, $71.5\%\}$, $\{71.5\%$, $71.5\%\}$, $\{71.9\%$, $72.0\%\}$, and $\{71.4\%$, $71.7\%\}$, producing an average set as $\{71.6_{\pm0.2}\%$, $71.7_{\pm0.3}\%\}$. AMContrast3D++ achieves a $mean$ of $71.7\%$, which performs better than many competitors. It also attains improvements against the baseline that performs better than PointNeXt by $1.7\%$ and AMContrast3D by $0.5\%$ in terms of $mean$ on average at test time. Meanwhile, AMContrast3D++ decreases $std$ ($\pm 1.0 \rightarrow \pm 0.3$), which means all results tend to be close to the $mean$, indicating that it outperforms our previous method with enhanced robustness.
 
We also perform the comparisons regarding module components and computation costs for PointNeXt, AMContrast3D, and AMContrast3D++. Table~\ref{tab:compare} thoroughly shows that AMContrast3D integrates the modules AEF and MG with the objective $\mathcal{L}_{AM}^s$, which obtains mIoU gains without any improvement on space complexity. AMContrast3D++ has slight increases in the model size ($41.59$M $\rightarrow$ $41.63$M) and the FLOPs ($364.57$G $\rightarrow$ $365.20$G) by combining APM, MR, and $\mathcal{L}_{REG}^s$. AMContrast3D++ introduces parameter counts, resulting in an additional time complexity in training hours, particularly during the forward pass to process extra computations. The additional parameters enable the model to capture enhanced representations. In summary, our methods, AMContrast3D and AMContrast3D++, effectively achieve better results.

\subsubsection{Qualitative Visualization}
\begin{figure*}[!t]
\centering
\includegraphics[width=6.8in]{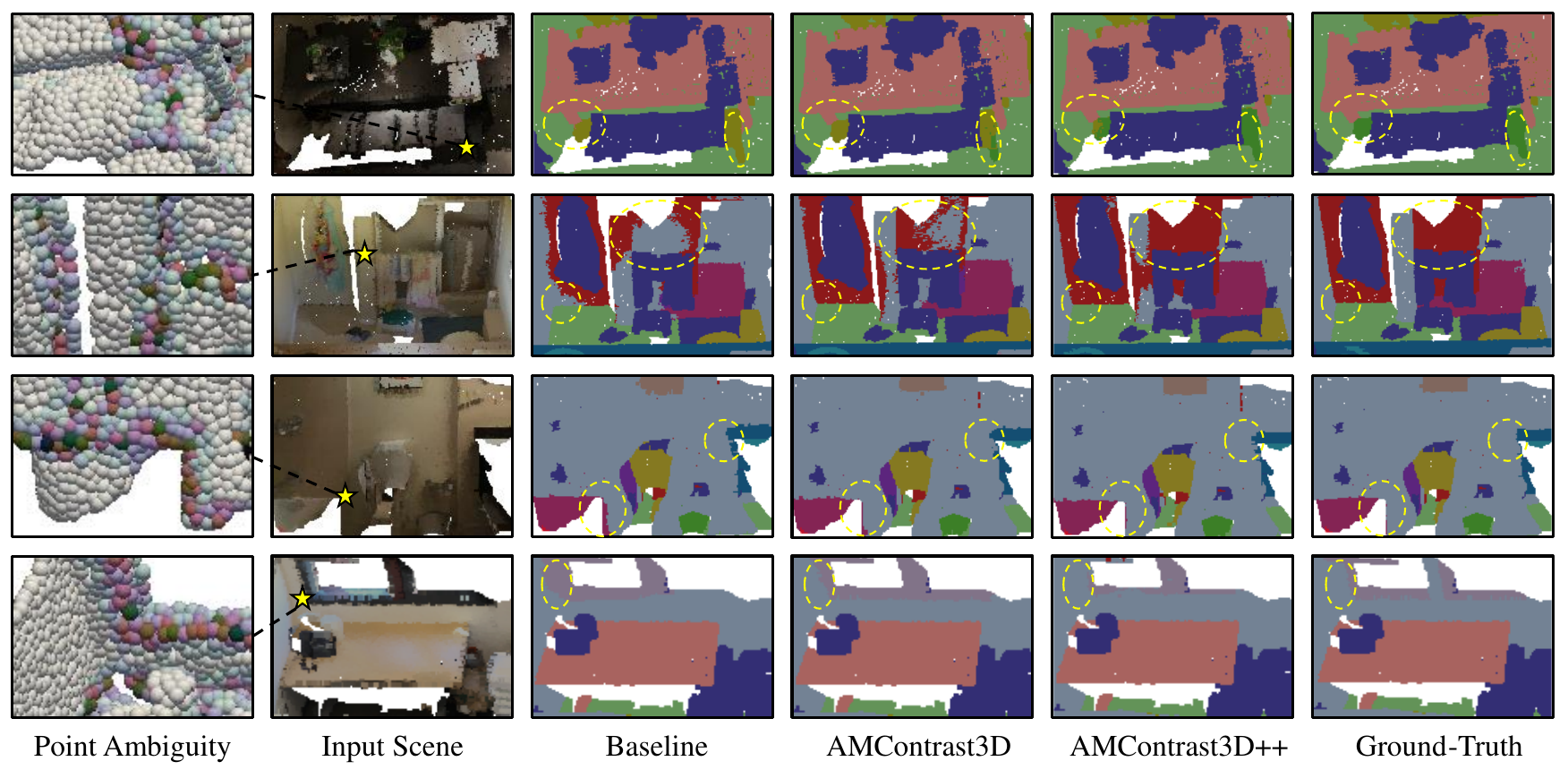}
 \caption{Visualization results on ScanNet dataset. We visualize the input scenes, the results predicted by the baseline (PointNeXt),  the results predicted by our method (AMContrast3D and AMContrast3D++), and the ground-truth semantic labels, where the yellow dashed circles indicate the gradual improvements. The yellow stars in the input scenes indicate the representative transition regions connecting different semantic classes, from which we zoom these regions and visualize the detailed ambiguity levels for individual points.}
\label{fig:scannet}
\end{figure*}

To provide an intuitive comparison, we visualize semantic segmentation results on challenging scenes. Fig.~\ref{fig:s3dis} and Fig.~\ref{fig:scannet} present the performances of S3DIS and ScanNet datasets. Starting from the leftmost figure and moving to the right, it shows the visualizations of point ambiguity, input scene, baseline, AMContrast3D, AMContrast3D++, and ground-truth results. We use the yellow dashed circles to highlight our improvements. Compared with the baseline, our AMContrast3D performs better, and AMContrast3D++ further outperforms the previous AMContrast3D with a leading performance, which also gradually aligns with the ground-truth result. These visual improvements are gained in the transition regions connecting several semantic classes. Taking the real input scene as a reference, we use yellow stars to locate the transition regions, where the point ambiguity stems from these regions. Thereafter, we zoom these regions and use a color map to assign ambiguities in the first columns, where Fig.~\ref{fig:ambiguity} in Section~\ref{sec:method_AAA_AEF} presents the detailed color map ranging from low level $0$ to high level $1$. Improvements in transition regions validate the effectiveness of our methods. In terms of quantitative and qualitative comparisons, our methods achieve better or more comparative results on challenging benchmarks through concise and straightforward approaches.

\subsection{Ablation Study}
\label{sec:exp_C_ablation}

\begin{table}[t]
\begin{center}
\setlength\tabcolsep{5.2pt} 
\caption{Ablation on temperature in controlling the contrast level. We highlight the best \colorbox{blue!8}{OA ($\%$)}, \colorbox{green!10}{mACC ($\%$)}, and \colorbox{lime!12}{mIoU ($\%$)}.} \label{tab:ablation_temper}
\begin{tabular}{c|c|c|c|c|c|c|c}
  \toprule
  Temperature $\tau$ & 0.1 & 0.2 & 0.3 & 0.4 & 0.5 & 0.6 & 0.7 \\
  \midrule
  OA   & 90.6 & 90.7 & \cellcolor{blue!8} 91.1 & 91.0 & 90.7 & 90.6 & 89.8\\
  mACC & \cellcolor{green!10} 77.6 & 76.4 & 77.1 & 77.4 & 75.8 & 76.5 & 73.5\\
  mIoU & 70.2 & 70.1 & \cellcolor{lime!12} 71.8 & 71.3 & 69.2 & 70.8  & 66.9\\
  \midrule
  Parameter $\beta$ & \multicolumn{2}{c|}{0.008} & \multicolumn{2}{c|}{0.04} & \multicolumn{2}{c|}{0.1} & 2.0 \\
  \midrule
  OA   & \multicolumn{2}{c|}{90.9} & \multicolumn{2}{c|}{\cellcolor{blue!8} 91.1} & \multicolumn{2}{c|}{90.7} & 90.2 \\
  mACC & \multicolumn{2}{c|}{77.2} & \multicolumn{2}{c|}{77.1} & \multicolumn{2}{c|}{77.3} & \cellcolor{green!10} 77.9 \\
  mIoU & \multicolumn{2}{c|}{70.7} & \multicolumn{2}{c|}{\cellcolor{lime!12} 71.8} & \multicolumn{2}{c|}{71.3} & 70.0 \\
  \bottomrule
\end{tabular}
\end{center}
\end{table}

We further design ablations to demonstrate the performance of each method. To study the key modules in the ambiguity awareness branch, we mainly investigate effects with various combinations of design choices. All experiments are conducted on the S3DIS dataset.

\subsubsection{Temperature \texorpdfstring{$\tau$}{Lg} in Contrastive Learning and Parameter \texorpdfstring{$\beta$}{Lg} in Closeness Centrality} 

Contrastive learning improves training regularization, and the temperature $\tau$ scales the sensitivity of contrast levels. We validate its effects by setting different $\tau$ in~(\ref{eq:EMB_mi}). As shown in Table~\ref{tab:ablation_temper}, we find that a proper $\tau$ is within a range from $0.1$ to $0.7$, and the best result is achieved when $\tau=0.3$ for generating contrastive embeddings. A small $\tau$ accentuates the contrast level when the feature embeddings are different, resulting in a large gradient. Instead, a large $\tau$ is more forgiving of the embedding difference, which needs more training epochs to intensify the contrast effects and gain competitive results. Besides, $\beta$ is a tuning parameter to control the point-level ambiguity estimation, where it influences the range of closeness centrality. A larger $\beta$ results in a narrower range, and it is reasonable to have some fluctuations when the range changes. We set $\beta$ to be $0.04$ for its highest performance based on other experimental settings, including $0.008$, $0.1$, and $2.0$.

\subsubsection{Neighboring Size \texorpdfstring{$K$}{Lg} in AEF and \texorpdfstring{$\widetilde{K}$}{Lg} in MR} 

Neighboring sizes are used to aggregate local information. We first conduct analysis for neighboring size $K$ as a constraint to explore the intra-set and inter-set in~(\ref{eq:A}), constructing AEF with $K=6, 12, 24$. Next, we generally fix $K=24$ for the AMContrast3D approach and investigate $\widetilde{K}$ in MR for the AMContrast3D++ approach. $\widetilde{K}$ defines a group of points for refinement in~(\ref{eq:CrossMask}), which limits the searching boundary, thereby requiring a small size with more precise refining. We evaluate $\widetilde{K}=8, 12, 16$. Table~\ref{tab:ablation_neighbor} indicates that insufficient or excessive neighboring size affects the capability to describe local structural information of 3D point clouds, and a moderate neighboring size leads to better results.

\begin{table}[t]
\begin{center}
\caption{Ablation on neighboring sizes of AEF and MR modules.} \label{tab:ablation_neighbor}
\begin{tabular}{c|c|c|c|c|c|c|c}
  \toprule
  \multicolumn{4}{c|}{AEF} & \multicolumn{4}{c}{MR}\\
  \midrule
  $K$ &  OA & mACC & mIoU & $\widetilde{K}$ &  OA & mACC & mIoU 
  \\
  \midrule
  6  & 90.9 & \cellcolor{green!10} 77.2 & 70.5 & 8 & 90.5 & 76.4 & 69.7 \\
  12 & 90.6 & 76.0 & 69.8 & 12 & \cellcolor{blue!8} 91.2 & \cellcolor{green!10} 77.8 & \cellcolor{lime!12} 71.7 \\
  24 & \cellcolor{blue!8} 91.1 & 77.1 & \cellcolor{lime!12} 71.8 & 16 & 91.0 & 77.4 & 70.9\\
  \bottomrule
\end{tabular}
\end{center}
\end{table}

\begin{table}[t]
\begin{center}
\caption{Ablation on constant and adaptive margins.}
\label{tab:ablation_margin}
\begin{tabular}{c|c|c|c|c|c}
  \toprule
  Margin & Value & $\mu$ & $\nu$ & $m_i$ & mIoU
  \\
  \midrule
  Constant & $\bigcirc$ & $0$ & $0$ & $0$  & 70.5 \\
  Constant & $\uparrow$ & $0$ & $0.5$ & $0.5$ & 69.7 \\
  \midrule
  Adaptive & $\uparrow$ & $1$ & $0$ & $a_i$ & 70.1 \\
  Adaptive & $\uparrow$ $\bigcirc$ & $-1$ & $1$ & $1 - a_i$ & 70.6 \\
  Adaptive & $\uparrow$ $\bigcirc$ $\downarrow$ & $-1$ & $0.5$ & $0.5 - a_i$ & \cellcolor{lime!12} 71.8 \\
  Adaptive & $\uparrow$ $\bigcirc$ & $-1$ & $0.5$ & $max(0, 0.5 - a_i)$ & 70.5 \\
  \bottomrule
\end{tabular}
\end{center}
\end{table}

\begin{table}[t]
\begin{center}
\caption{Ablation on the percentage of point amounts on each ambiguity level with various neighboring sizes. We highlight the percentages for \colorbox{gray!16}{ambiguous points}, where the top percentage and the second-top percentage are \textbf{bolded} and \underline{underlined}, respectively.  Note that the sum within each neighboring size is $100\%$.
}
\label{tab:ablation_ambiguity}
\begin{tabular}{l|c|c|c|c}
  \toprule
  \makecell*[c]{\multirow{2}*{\makecell[c]{Ambiguity \\ level}}} & \multicolumn{4}{c}{Neighboring size $K$}  \\
  \cmidrule(lr){2-5}
   & $12$ & $18$ & $24$ & $30$ \\
  \midrule
  unambiguous: $a_i = 0$                & 98.37$\%$ & 97.11$\%$ &  96.03$\%$ & 95.25$\%$ \\
  \midrule
  \rowcolor{gray!16} low-$a_i$: $a_i \in (0,0.5)$   & \textbf{1.10}$\%$ & \textbf{2.34}$\%$ &  \textbf{3.41}$\%$ & \textbf{4.18}$\%$ \\
  \rowcolor{gray!16} semi-$a_i$: $a_i = 0.5$        & 0.00$\%$ & 0.01$\%$ &  0.01$\%$ & 0.02$\%$ \\
 \rowcolor{gray!16}  high-$a_i$: $a_i \in (0.5,1)$  & 0.02$\%$ & 0.03$\%$ &  0.05$\%$ &0.07$\%$ \\
  \rowcolor{gray!16} high-$a_i$: $a_i = 1$          & \underline{0.51}$\%$ & \underline{0.51}$\%$ &  \underline{0.50}$\%$ & \underline{0.48}$\%$\\
  \midrule
  OA     & 90.6 & 90.9 & \cellcolor{blue!8} 91.1 & 90.7 \\
  mACC   & 76.0 & 76.4 & \cellcolor{green!10} 77.1 & 76.7 \\
  mIoU    & 69.8 & 70.4 & \cellcolor{lime!12} 71.8 &  70.6  \\
  \bottomrule
\end{tabular}
\end{center}
\end{table}

\begin{table}[t]
\begin{center}
\caption{Ablation on the threshold and refining rate in the masked refinement mechanism. Note that the thresholds and refining rates are within a range between $0$ and $1$.} \label{tab:ablation_MR}
\begin{tabular}{c|c|c|c|c|c|c}
  \toprule
  \makecell[c]{Threshold \\ $[\varepsilon, \dot{\varepsilon}]$} &  \makecell[c]{OA \\ (Val)} & \makecell[c]{mACC \\ (Val)} & \makecell[c]{mIoU \\ (Val)} &  \makecell[c]{OA \\ (Test)} & \makecell[c]{mACC \\ (Test)} & \makecell[c]{mIoU \\ (Test)} \\
  \midrule
  $[0.88, 1.00]$  & 89.9  & 76.8  & 70.0  & 90.8 & 77.4	& 71.1 \\
  $[0.90, 1.00]$ & \cellcolor{blue!8} 90.4 & \cellcolor{green!10} 77.5	& \cellcolor{lime!12} 70.6	& \cellcolor{blue!8} 91.2	& \cellcolor{green!10} 77.8	& \cellcolor{lime!12} 71.7 \\
  $[0.92, 1.00]$ & 89.9 & 76.9	& 69.9	& 90.7	& 77.4	& 70.9 \\
  $[0.70, 0.94]$ & 90.3 & 77.1 & 70.4 & 90.3 & 77.5 & 70.5 \\
  $[0.61, 0.82]$ & 90.0 & 76.7 & 69.7 & 90.1 & 77.1 & 69.9 \\
  \midrule
  \makecell[c]{Refining \\ rate $\gamma$} &  \makecell[c]{OA \\ (Val)} & \makecell[c]{mACC \\ (Val)} & \makecell[c]{mIoU \\ (Val)} &  \makecell[c]{OA \\ (Test)} & \makecell[c]{mACC \\ (Test)} & \makecell[c]{mIoU \\ (Test)} \\
  \midrule
  0.2 & 90.3	& \cellcolor{green!10} 77.5	& 70.5	& 91.1	& \cellcolor{green!10} 78.0	& 71.4 \\
  0.4 & 90.2	& 76.7	& 70.2	& 91.0	& 77.0	& 71.0 \\
  0.9 & 90.2	& 76.2	& 69.9	& 91.0	& 77.0	& 71.2 \\
  1.0 & \cellcolor{blue!8} 90.4 & \cellcolor{green!10} 77.5	& \cellcolor{lime!12} 70.6	& \cellcolor{blue!8} 91.2	&  77.8	& \cellcolor{lime!12} 71.7 \\
  \bottomrule
\end{tabular}
\end{center}
\end{table}

\subsubsection{Adaptive Margin \texorpdfstring{$m_i$}{Lg} in MG} 

To evaluate the effects of margin $m_i=\mu \cdot a_i+ \nu$, we conduct an ablation with different settings in Table~\ref{tab:ablation_margin}. The first two rows are constant margins with the remaining rows being adaptive margins. Constant $m_i$ generates uniform contrastive objectives without considering the disparity of ambiguities, which may not capture sufficient context for high-ambiguity points, and the performance significantly drops. The best mIoU result is achieved with $\mu=-1$ and $\nu=0.5$ that uses adaptive $m_i$ controlled by $a_i \in (0,1]$ in~(\ref{eq:AMG_mi}). In this case, margins cover positive ($\uparrow$), zero ($\bigcirc$), and negative ($\downarrow$) values to determine the expansion or shrinkage of decision boundaries. This ablation suggests that negative values are essential for ambiguity-aware margins to boost the segmentation performance.

\subsubsection{Percentage of Point Amounts on each Ambiguity Level \texorpdfstring{$a_i$}{Lg}}

We categorize unambiguous points, low-ambiguity points, semi-ambiguity points, and high-ambiguity points, analyzing the percentage of point amounts on each ambiguity level $a_i$ by validating several neighboring sizes $K$ in Table~\ref{tab:ablation_ambiguity}. In general, ambiguous points take small amounts compared to unambiguous points since transition regions are fewer than inner regions in 3D space. For ambiguous points, points with $a_i \in (0,0.5)$ account for the largest percentages, indicating that most points are with low ambiguities. Instead, the lowest percentages are semi-ambiguity points without a discrepancy between two kinds of closeness centrality in~(\ref{eq:G}), which is rare because of the irregularity and sparsity of points. As delineated in Section~\ref{sec:method_AAA_AEF}, the size, denoted as $K$, determines the amount of local points. We observe that when $K$ is enlarged, more ambiguous points are involved, attracting ambiguity awareness through our methods. The leading results are with $K=24$, and ambiguous points have reasonable amounts, which are $3.41\%$, $0.01\%$, and $0.55\%$ for low, semi, and high ambiguities, respectively. Thus, our contrastive regularization is adaptive on different ambiguity levels for individual points.

\subsubsection{Threshold of \texorpdfstring{$\varepsilon, \dot{\varepsilon}$}{Lg} and Refining Rate \texorpdfstring{$\gamma$}{Lg} in MR}

We study the effects of MR for AMContrast3D++, which contains a threshold $[\varepsilon, \dot{\varepsilon}]$ in~(\ref{eq:SelfMask}) and a rate $\gamma$ to control the refinement in~(\ref{eq:EmbRefine_3}). We define $\varepsilon$ as a lower bound of ambiguity levels, where the upper bound $\dot{\varepsilon}$ is fixed as $1$ because of its extremely high ambiguity. We first evaluate different settings of $\varepsilon$ as $0.88$, $0.90$, and $0.92$. When we decrease the upper bound $\dot{\varepsilon}$, \textit{e.g.}, $0.94$ and $0.82$, the corresponding lower bound $\varepsilon$ might be changed since $\varepsilon$ is always less than or equal to $\dot{\varepsilon}$, and we observe that the results of small $\dot{\varepsilon}$ are not as good as large $\dot{\varepsilon}$. As shown in Table~\ref{tab:ablation_MR}, the best results are achieved when $\varepsilon$ is $0.9$ and $\gamma$ is $1$, where MR conducts full refinement for points with high-ambiguity levels between $0.9$ and $1$. Moreover, MR does not cause significant fluctuations in different settings, indicating that the proposed MR is not sensitive to the hyperparameters, and our method has stability.

\subsection{Discussion}
\label{sec:exp_D_discussion}

\subsubsection{Technical Novelty}
\label{sec:disc_A_novelty}

We introduce contrastive learning as a comprehensive ambiguity-aware regularization through joint training in AMContrast3D++ based on its preliminary version, AMContrast3D. Instead of simple combinations and increments of existing methods, we highlight that our approach is grounded in three key innovations: the definition, estimation, and exploitation of point ambiguity in 3D point clouds. \textit{First}, unlike prior approaches that rely on global or local levels, we focus on individual point levels by defining ambiguity as a measurable property of each point. This definition forms the foundation for the ambiguity to be explicitly addressed, providing a new perspective to reflect inherent uncertainty among points. \textit{Second}, we design an estimation framework to dynamically quantify point-level ambiguities. Derived from rigorous mathematical formulations, the estimated ambiguities enable the semantic segmentation branch equipped to handle challenging transition regions with diverse semantics. \textit{Third}, the exploitation lies in optimization and refinement processes. We optimize contrastive learning by adaptive margin confidences and predict ambiguities to refine semantic embeddings. Through these innovations, our approach addresses the challenges of ambiguity in 3D point cloud segmentation. Experiments validate that AMContrast3D and AMContrast3D++ contribute to enhanced robustness and refined segmentation, demonstrating the effectiveness of our definition, estimation, and exploitation of ambiguity.

\subsubsection{Potential Application and Limitation}
\label{sec:disc_B_background}

As our approaches are general ambiguity-aware embedding learning frameworks, they can be extended to various point cloud applications in 3D environments. For example, indoor service robots enable efficient navigation and obstacle avoidance within surroundings such as homes, offices, and warehouses. By providing semantic perception with ambiguity awareness, our frameworks may assist robots in interpreting complex objects of transitional regions, such as a table near a wall and a chair positioned around a doorway. Thus, the enhanced understanding allows for precise path planning through semantic comprehension of the environment, contributing to reliable navigation. Since we focus on indoor and static scenes like most prior methods, application scenarios have limitations on outdoor and dynamic scenes, such as autonomous driving applications. Such applications demand real-time tracking on moving objects and changing conditions, thereby involving spatial and temporal coherence. We aim to delve deeper on addressing temporal modeling challenges with ambiguity awareness to extend the applicability in various scenarios.

\section{Conclusion and Future Work}
\label{sec:conclu}

In this paper, we have introduced AMContrast3D, a framework that integrates point ambiguity with an ambiguity estimation framework and an adaptive margin generator in adaptive margin contrastive learning. To further harness the modeling potential of ambiguity awareness, we extend this method to AMContrast3D++, which incorporates an ambiguity prediction module and a masked refinement mechanism. Our method jointly predicts per-point ambiguities and semantic labels, enabling the refinement of learnable embeddings in a comprehensive approach. Extensive experiments and visualizations demonstrate the superior performance of the proposed methods in 3D point cloud segmentation.

We hope our methods inspire a new perspective on 3D ambiguity awareness. Considering the time-intensive nature of point labeling, our methods hold potential for transitioning from fully to weakly supervised settings. Per-point ambiguity estimation may advance the pseudo-label generation, addressing the challenges associated with incomplete annotations in large-scale datasets and realizing more effective segmentation. In addition, a promising extension lies in other 3D tasks beyond point cloud tasks. 3D Gaussian provides better representations than a point cloud by modeling the distributions in 3D space, which allows for adapting point cloud pipelines to Gaussian-based approaches, facilitating tasks such as 3D reconstruction and novel view synthesis. Furthermore, integrating ambiguity awareness into Gaussian splatting has the potential to enhance spatial uncertainty and temporal variability in 3D perception.


%


\bibliographystyle{IEEEtran}
\bibliography{my_ref}



 




\vfill

\end{document}